\documentclass{article}

\usepackage{arxiv}

\usepackage[utf8]{inputenc} 
\usepackage[T1]{fontenc}    
\usepackage{url}            
\usepackage{amsfonts}       
\usepackage{nicefrac}       
\usepackage{microtype}      
\usepackage{lipsum}
\usepackage{graphicx}
\usepackage{booktabs}
\usepackage{tabularx}
\usepackage{siunitx}
\usepackage{placeins}
\usepackage{soul}
\usepackage{multirow}
\usepackage{mathtools}
\usepackage{pgfplots}
\usepackage{colortbl}
\usepackage{xcolor}
\usepackage{wrapfig}
\usepackage{makecell}

\usepgfplotslibrary{groupplots}
\usepackage{subcaption}

\usepackage{xcolor} 
\usepackage[normalem]{ulem}
\usepackage[accsupp]{axessibility}  
\usepackage{hyperref}
\usepackage{orcidlink}
\usepackage{cleveref}
\usepackage{svg}
\newcommand{\up}{$\uparrow$}
\newcommand{\down}{$\downarrow$}

\title{SynSur: An end-to-end generative pipeline for synthetic industrial surface defect generation and detection}

\author{
Paul Julius Kühn\orcidlink{0000-0003-2458-0030} \\
Fraunhofer IGD, 64283 Darmstadt, Germany \And
Mika Pommeranz\orcidlink{0009-0009-8727-8235} \\ 
Fraunhofer IGD, 64283 Darmstadt, Germany \And
Arjan Kuijper\orcidlink{0000-0002-6413-0061} \\
Fraunhofer IGD, 64283 Darmstadt, Germany \And
Saptarshi Neil Sinha\orcidlink{0000-0001-6637-0379} \\
Fraunhofer IGD, 64283 Darmstadt, Germany
}

\begin{document}
\maketitle
\begin{abstract}
Industrial surface defect inspection suffers from a fundamental data bottleneck: defects are rare, annotations require expert knowledge, and collecting balanced training sets is slow and costly. We present SynSur, an end-to-end pipeline for synthetic defect generation and automatic annotation, designed to reduce the manual effort and data scarcity that limit deployed inspection systems. The pipeline combines Vision-Language-Model–based prompt construction, LoRA-adapted diffusion, mask-guided inpainting, metric-based sample filtering, and automatic label derivation. We evaluate SynSur on BSData (pitting defects on ball screw drives) and the scratch subset of MSD, reporting downstream detection performance across YOLOX, YOLOv26, and LW-DETR under real-only, synthetic-only, mixed, and union training regimes. The full generation pipeline processes a candidate pool of 1,000 images and delivers a filtered, annotated synthetic dataset with no manual labeling effort. Synthetic-only training does not replace real data; however, augmenting the full real set with synthetic samples yields consistent AP gains in selected configurations, and augmenting scarce real sets substantially reduces the performance gap to full real-data training. The cross-domain study on MSD confirms pipeline portability while highlighting the importance of domain-specific adaptation and annotation quality control. SynSur demonstrates that diffusion-based defect synthesis is a practically viable tool for strengthening industrial inspection pipelines, particularly where labeled data collection is the primary bottleneck.

\end{abstract}
\section{Introduction}\label{sec:introduction}
Deep learning has significantly advanced visual defect detection in industrial quality 
assurance~\cite{wang2025review}. 
However, these methods require large amounts of labeled data, 
which is difficult to obtain in practice: defects occur rarely, 
and annotation demands expert knowledge. 
The resulting small, imbalanced datasets 
limit model robustness and generalization~\cite{jin2022survey,wang2025review}.
Synthetic data generation offers a promising solution. 
Traditional approaches rely on 3D rendering pipelines~\cite{denninger2020blenderproc,tremblay2018training,tremblay2018falling,sinha20256dstrawberryposeestimation} 
that provide explicit control over defect 
geometry and automatic labeling, but require detailed scene models and substantial manual 
effort~\cite{bai2025comprehensivesurveymachinelearning}. 
Moreover, the sim-to-real gap--caused by differences in texture, 
lighting, and sensor characteristics--often degrades detection performance on real data~\cite{wang2025review}.
The core industrial challenge is clear: annotation budgets are fixed, defects are rare, and waiting for sufficient real data before deploying a detector is not operationally viable. SynSur addresses this by providing a fully automated path from a small set of real defect images to a filtered, annotated synthetic training set — requiring no manual prompt engineering, no manual sample curation, and no manual labeling beyond the seed annotations already needed to train any supervised detector.
Hence in this work, we propose an end-to-end pipeline that derives an initial defect prompt from a Vision-Language Model (VLM), 
adapts a diffusion backbone with LoRA~\cite{hu2021loralowrankadaptationlarge}, 
inserts defects by mask-guided inpainting, filters the generated patches, 
and derives detection labels automatically. Furthermore, this enables us to evaluate how far diffusion-based models can be used as a 
practical source of training data when real annotations are scarce.
We evaluate the pipeline on BSData~\cite{schlagenhauf2021industrial}, 
an industrial dataset of pitting defects on ball screw drives, 
analyze both the visual quality of the generated defects and their utility for downstream detection,
and extend the study with a cross-dataset evaluation on the scratch subset of Mobile Phone Screen Surface Defect Dataset
(MSD)~\cite{zhang2022fdsnet} to assess which stages transfer to a second defect domain.
Our main contributions, each designed with automation and scalability as first-order constraints, are:
\begin{itemize}
    \item \textbf{SynSur:} An end-to-end pipeline for industrial defect synthesis and annotation that combines VLM-based prompt generation, 
    LoRA-adapted diffusion, mask-guided inpainting, sample filtering, and automatic label derivation for detector training. To enable reproducibility, we will release the synthetic defect generation code and all data used in our evaluations upon acceptance of the paper.
    \item \textbf{Systematic ablation:} A systematic analysis of prompt construction, LoRA selection, and synthetic sample filtering on BSData~\cite{schlagenhauf2021industrial}, 
    linking synthetic defect quality to downstream detection utility.
    \item \textbf{Evaluation \& domain transfer:} We evaluated our approach against baseline detectors that employ both CNN- and transformer-based architectures across combinations of real-only, synthetic-only, mixed, and union training regimes. Our results show that synthetic data is most effective as augmentation when real data is scarce.
\end{itemize}

\section{Related Work}\label{sec:related}

\textbf{Deep learning methods.} dominate industrial defect detection.
One-stage detectors such as YOLO~\cite{redmon2016yolo} and its
variants~\cite{redmon2018yolov3,bochkovskiy2020yolov4,sapkota2026yolo26keyarchitecturalenhancements}
remain attractive for real-time inspection, while transformer-based models derived from
DETR~\cite{carion2020detr,lv2023detrs,lv2024rtdetrv2improvedbaselinebagoffreebies,chen2024lwdetrtransformerreplacementyolo,rf-detr}
offer an alternative set of inductive biases. Supervised detection performance depends
strongly on data quantity and class
balance~\cite{bai2025comprehensivesurveymachinelearning,wang2025review}, motivating
synthetic data generation when annotated industrial defects are scarce. 3D rendering
pipelines address this by modeling geometry, materials, and illumination with automatic
labels~\cite{xu2024systematic}, but simulation still requires detailed scene models and
the remaining domain gap can limit transfer to real sensor data.\\
\noindent\textbf{Image-based synthesis.} creates defects through copy-paste, patch insertion, or procedural
perturbation~\cite{jain2022synthetic,lebert2022synthetic}. CutPaste~\cite{li2021cutpaste}
and DRAEM~\cite{zavrtanik2021draem} showed that synthetic anomalies can support anomaly
detection even without real defect samples. GAN-based approaches, including
pix2pix~\cite{isola2017pix2pix}, CycleGAN~\cite{zhu2017cyclegan}, and
DefectGAN~\cite{liu2024defectgan}, extend this idea to learned defect generation, though
defect insertion and defect--background interaction are often not realistic enough for
reliable supervised detection gains~\cite{nguyen2025we}.\\
\noindent\textbf{Generative Diffusion models.}~\cite{ho2020ddpm,rombach2022stablediffusion} have become strong
generative backbones and often surpass GANs in visual
fidelity~\cite{dhariwal2021diffusionbeatsgans}. Control mechanisms such as
ControlNet~\cite{zhang2023controlnet} and personalization methods such as
DreamBooth~\cite{ruiz2023dreambooth} and
LoRA~\cite{hu2021loralowrankadaptationlarge} make them suitable for domain-specific
synthesis from limited data. Recent industrial inspection work uses inpainting and
few-shot diffusion for realistic defect
insertion~\cite{song2025defectfill,tayeb2024defectgen,tayeb2025defectdiffusion,ali2024anomalycontrol},
yet downstream value still depends strongly on prompt quality, spatial control, and
sample selection~\cite{azizi2023synthetic,nguyen2025we,shipard2023diversity}. VLMs such
as BLIP-2~\cite{li2023blip2}, LLaVA~\cite{liu2024llava}, and
Qwen2-VL~\cite{wang2024qwen2vl} can automatically extract domain-specific image
attributes to replace manual prompt engineering in diffusion-based synthesis. Although
widely applied to captioning and retrieval, their use for industrial defect prompt
construction remains underexplored.\\
Existing work typically emphasizes individual aspects such as realism, few-shot
generation, or augmentation effects~\cite{jain2022synthetic,lebert2022synthetic,li2021cutpaste}. Our focus is to investigate these components jointly by integrating VLM-based prompt construction, mask-guided inpainting, LoRA-based domain adaptation, quality filtering, and automatic annotation within a unified end-to-end pipeline for object detection training. Furthermore, we study the impact of our synthetically generated data on detection network performance in an industrial domain.

\section{Methodology}\label{sec:methodology}

\begin{figure}[htb!]
  \centering
  \includegraphics[width=1.\textwidth]{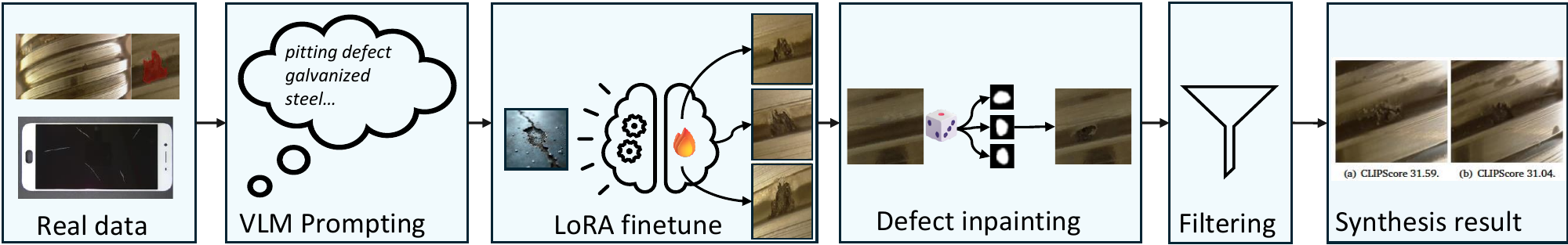}
  \caption{\textbf{Overview of SynSur:} End-to-end pipeline for synthetic defect data generation. From left to right: data preparation, VLM prompt construction, LoRA fine-tuning, random inpainting-based defect generation with defect segmentation,image blending and automated annotation. A metric filtering based on CLIPScore~\cite{hessel2021clipscore} DreamSim~\cite{fu2023dreamsim} leads to the final results and ensures plausible synthesis.}
  \label{fig:data_pipeline}
\end{figure}

Fig.~\ref{fig:data_pipeline} summarizes the end-to-end synthetic defect generation pipeline. 
We first extract defect patches for prompt construction and 
LoRA~\cite{hu2021loralowrankadaptationlarge} training. 
Synthetic masks derived from real annotations then define the 
inpainting regions for Flux.1-dev~\cite{flux2024,labs2025flux1kontextflowmatching} 
with the adapted LoRA~\cite{hu2021loralowrankadaptationlarge}. Candidate patches are then filtered with 
DreamSim~\cite{fu2023dreamsim} and CLIPScore~\cite{hessel2021clipscore} before SAM3~\cite{carion2025sam3segmentconcepts} 
segments the generated defects for blending and automatic annotation.

\subsection{Pipeline for synthetic defect generation}\label{subsec:arch}
\begin{figure}[ht]
    \centering
    \begin{subfigure}[b]{0.38\textwidth}
      \centering
      \includegraphics[width=\linewidth]{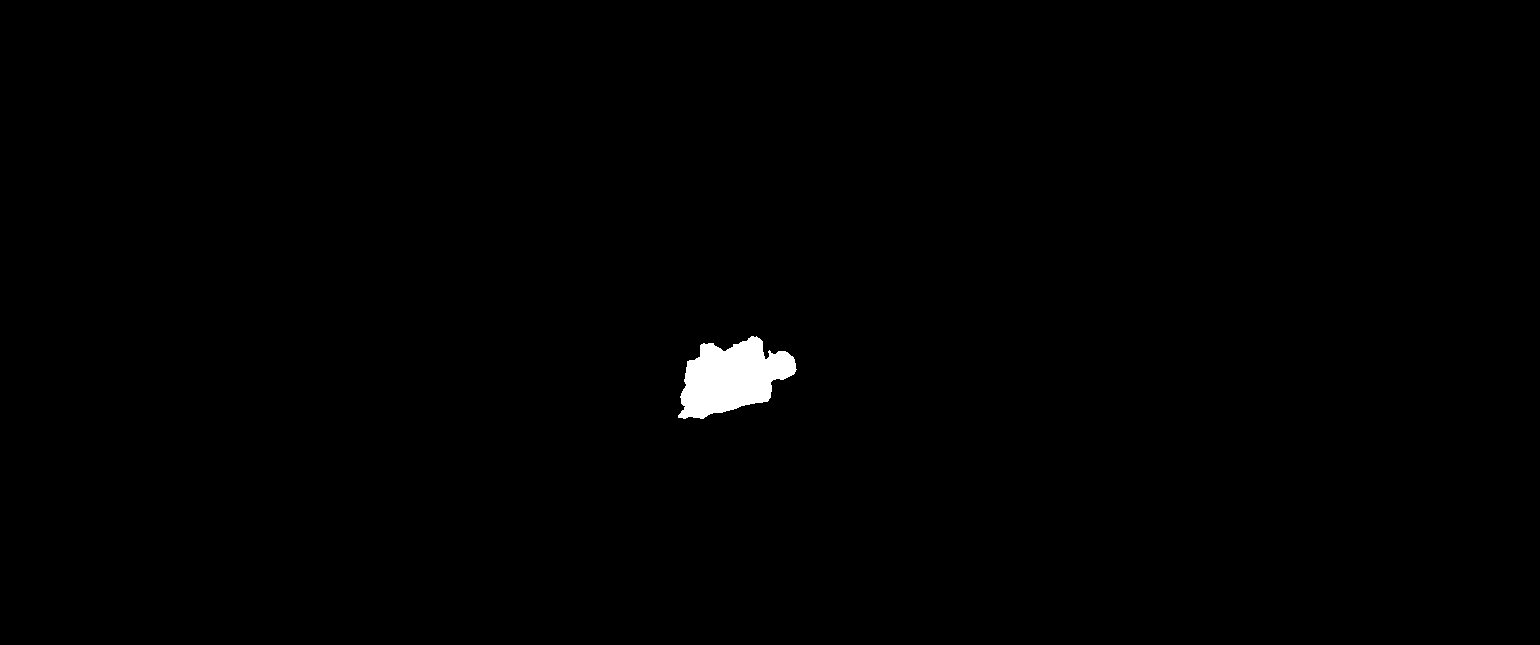}
      \caption{}
      \label{fig:synth_mask}
    \end{subfigure}
    \begin{subfigure}[b]{0.38\textwidth}
      \centering
      \includegraphics[width=\linewidth]{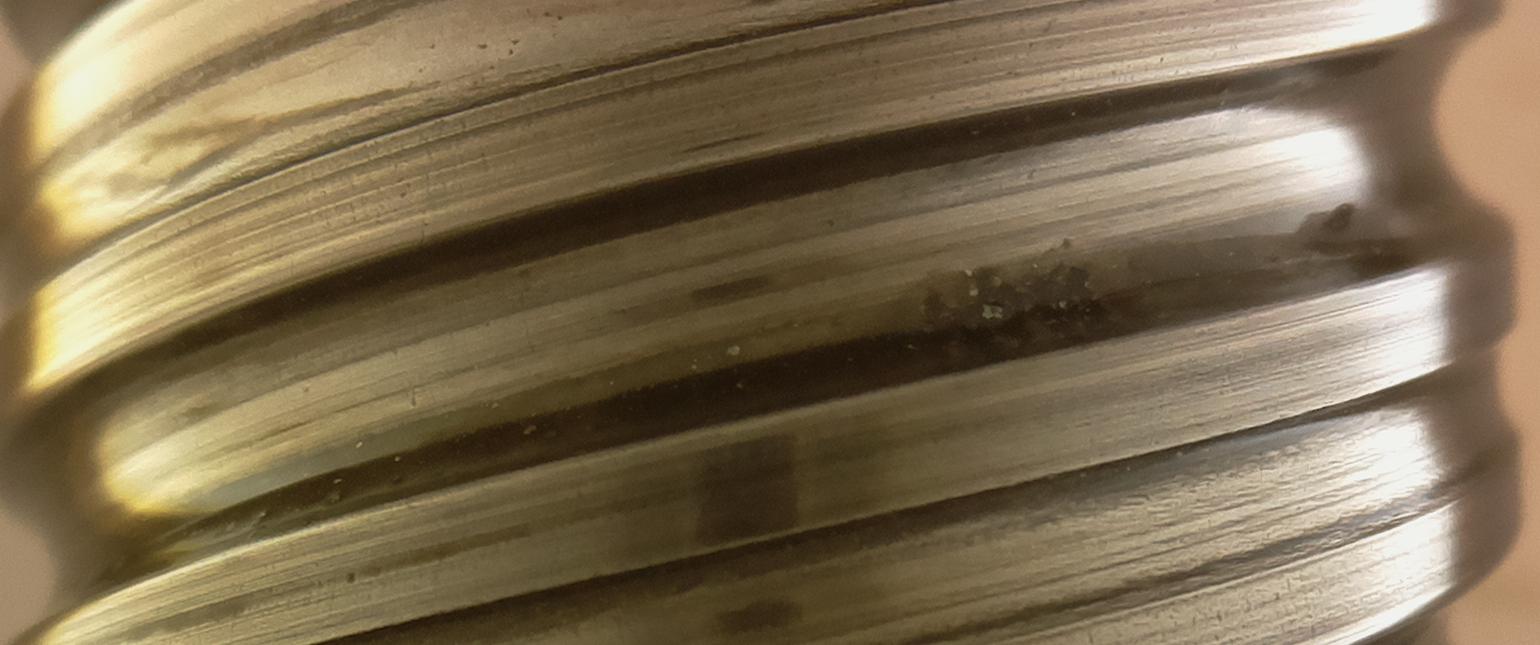}
      \caption{}
      \label{fig:defect_free}
    \end{subfigure}
    \begin{subfigure}[b]{0.16\textwidth}
      \centering
      \includegraphics[width=\linewidth]{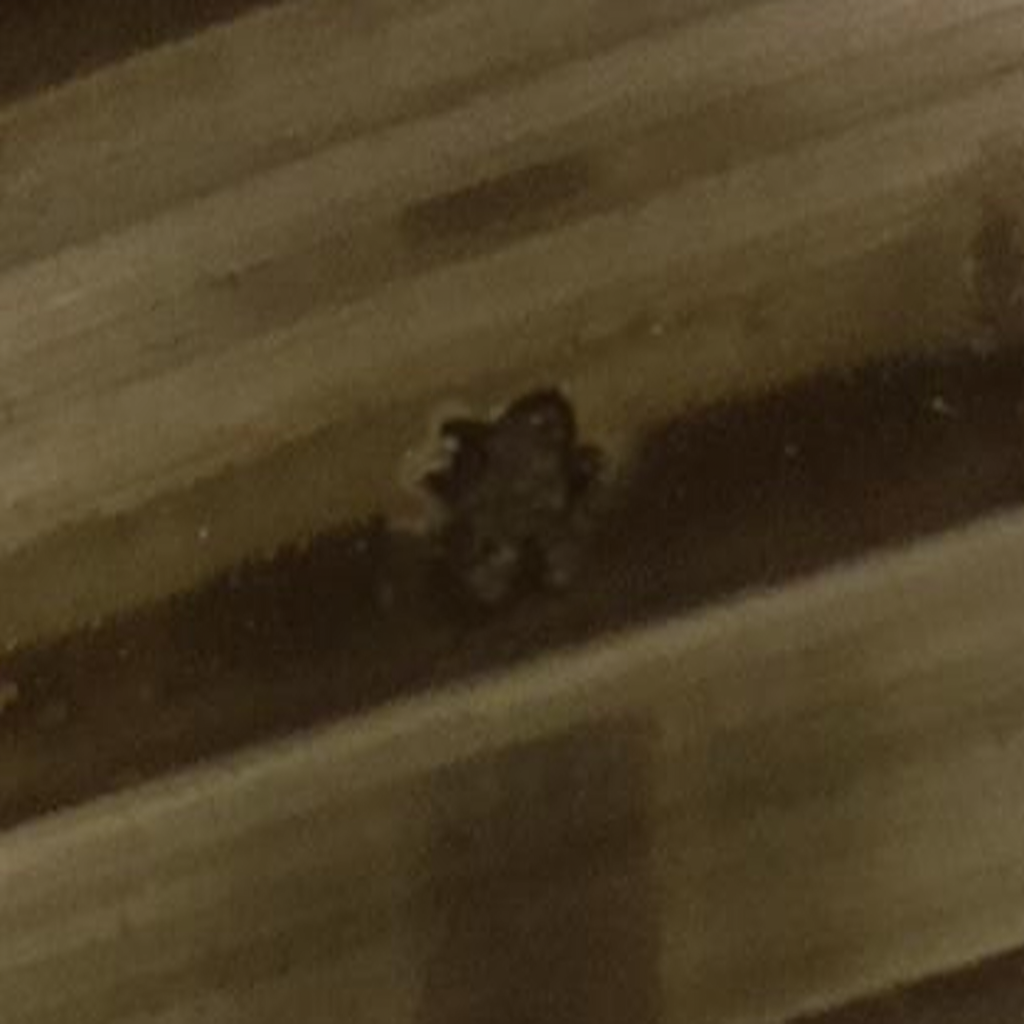}
      \caption{}
      \label{fig:generated_patch}
    \end{subfigure}
    \caption{Synthetic mask \textit{(a)} corresponding defect-free crop \textit{(b)}, 
    and defect patch \textit{(c)} generated from inputs \textit{(a)} and \textit{(b)}.}
    \label{fig:synth_patch}
\end{figure}

The individual steps of our synthetic defect generation pipeline (Fig.~\ref{fig:data_pipeline}) is presented below in detail:\\
\noindent\textbf{Prompt Generation.}
Text prompts guide the diffusion-based synthesis. We derive an initial prompt semi-automatically with 
Qwen2-VL~\cite{wang2024qwen2vl}, 
which processes batches of defect patches and returns 
structured tags describing material properties, 
defect appearance, and imaging conditions. 
The prompt is assembled from the most frequent tags after light pruning of overly generic descriptors 
and used for both LoRA~\cite{hu2021loralowrankadaptationlarge} training and inpainting.\\
\noindent
\textbf{Low Rank-Adaptation (LoRA).}
Prompts alone are insufficient for realistic defect synthesis because 
general-purpose diffusion models do not reliably capture industrial surface structure. 
We therefore adapt Flux.1-dev~\cite{flux2024} using LoRA~\cite{hu2021loralowrankadaptationlarge}. 
We compare two training strategies: (1)~\textit{random sampling}, 
which uses 100 randomly selected defect patches, and (2)~\textit{size-binned sampling}, 
which trains three separate LoRAs on small, medium, and large defects with 100 samples each. 
All variants use identical hyperparameters and train for 2{,}000 steps.
Model selection is based on DreamSim~\cite{fu2023dreamsim} and CLIPScore~\cite{hessel2021clipscore}, 
as detailed in Sec.~\ref{sec:synthdataqual}.\\
\noindent\textbf{Mask Generation and Inpainting.}
Inpainting masks define the synthesis region. 
We derive them from real training annotations 
to preserve realistic defect shapes. 
For additional spatial variation, masks are randomly scaled ($\times 0.8$--$1.2$) and shifted ($\pm 50$\,px). These ranges were chosen to remain within plausible defect-size variation observed in the BSData training set while avoiding mask placement at image boundaries, which would introduce inpainting artifacts at crop edges.
This yields 1{,}000 single-defect binary masks, 
with 500 masks for each image 
resolution (Fig.~\ref{fig:synth_mask}).
Synthetic defects are generated with 
Flux.1-dev~\cite{flux2024,labs2025flux1kontextflowmatching} 
and the selected LoRA~\cite{hu2021loralowrankadaptationlarge}.
For each sample, a defect-free image~\ref{fig:defect_free} is cropped around the mask 
region and resized to $1024{\times}1024$\,px with contextual 
padding of approximately twice the defect area. 
The model receives the background patch, 
mask, and prompt and generates a synthetic defect in 
30 sampling steps (Fig.~\ref{fig:generated_patch}).
We generate 1{,}000 candidate patches from varied defect-free images, masks, and random seeds. 
From this pool, we retain 420 patches, corresponding to approximately twice the number of real defect patches in the training split, 
based on DreamSim~\cite{fu2023dreamsim} and CLIPScore~\cite{hessel2021clipscore}. 
\\ 
\noindent\textbf{Label derivation.}
Since inpainting may introduce background artifacts, 
SAM~3~\cite{carion2025sam3segmentconcepts} refines the generated defect masks for BSData~\cite{schlagenhauf2021industrial} using the mask 
bounding box and a text cue. The resulting masks are smoothed with Gaussian blur for seamless blending into 
defect-free images. Annotations are generated in COCO~\cite{COCO} format.
End-to-end, the pipeline delivers 420 fully annotated synthetic images in approximately 8 GPU-hours on a single A100, with no manual labeling beyond seed annotations (details in Sec.~B in the supplementary material).
\FloatBarrier

\subsection{Training Setup}
We evaluate three detectors---YOLOX~\cite{ge2021yolox}, 
YOLOv26~\cite{sapkota2026yolo26keyarchitecturalenhancements}, 
and LW-DeTr~\cite{chen2024lwdetrtransformerreplacementyolo}—
under identical configurations to assess synthetic data utility.

\noindent\textbf{Detection Models.}
We evaluate YOLOX-S(9.0M)~\cite{ge2021yolox} and YOLOv26-S(9.5M)~\cite{sapkota2026yolo26keyarchitecturalenhancements} as 
one-stage CNN baselines, and LW-DETR-t(12M)~\cite{chen2024lwdetrtransformerreplacementyolo} as a 
lightweight transformer alternative.
All models use pretrained initialization from the COCO dataset~\cite{COCO}. 
Evaluating both architecture families allows assessing whether synthetic data benefits generalize across detector 
families with different inductive biases. 

\noindent\textbf{Training Configurations.}\label{sec:trainingconfig}
We compare real-only (100/0), synthetic-only (0/100), mixed (75/25, 50/50, 25/75), and union (100/100, 100/200) training 
configurations. Here, \emph{R/S} denotes the amount of real and synthetic training data relative to the full real training split, in \%.
Results are averaged over three runs with different random seeds (mean $\pm$ std).

\FloatBarrier

\section{Evaluation}
We evaluate the proposed pipeline along these dimensions: dataset
characteristics, synthetic defect quality by optimizing LoRA variant selection and prompt construction, downstream detection
performance, and cross-domain transferability. In order to evaluate these dimensions we perform both qualitative and quantitative analysis and provide a final qualitative analysis to summarize our results. We first characterize
BSData~\cite{schlagenhauf2021industrial}
scratch subset, then select the appropriate LoRA~\cite{hu2021loralowrankadaptationlarge} variant, analyze synthetic sample quality, detector
performance under varying real/synthetic data mixtures, and transfer to
a second industrial inspection domain utilizing the MSD dataset~\cite{zhang2022fdsnet}.
\subsection{Dataset analysis}
\noindent\textbf{BSData.}~\cite{schlagenhauf2021industrial} is an industrial surface defect 
dataset containing close-up images of ball screw drive spindles with pitting defects. 
The dataset comprises 1,104 images with 394 pixel-level segmentation masks captured under realistic 
laboratory conditions. Detection is challenging due to the small size and low contrast of defects. 
We filter the dataset to the two most frequent resolutions ($1130{\times}460$ and $1540{\times}645$ pixels) 
for consistent processing, resulting in 1,035 images: 710 defect-free and 325 defective, 
containing 357 annotated defect instances leading to a training data distribution of 65/15/20\%. 
Most defective images contain a single defect (299), 
while 20 contain two and 6 contain three instances. We analyze the spatial distribution of defects in the retained images (Figs.~10a and 10b in the supplementary material), which shows a strong bias toward the upper left quadrant, likely due to the recording setup. 
This non-uniform distribution motivates our mask generation
strategy in Sec.~A.1 in the supplementary material to ensure realistic defect placement.
\begin{figure}[ht]
        \centering
        \includegraphics[width=0.2\linewidth]{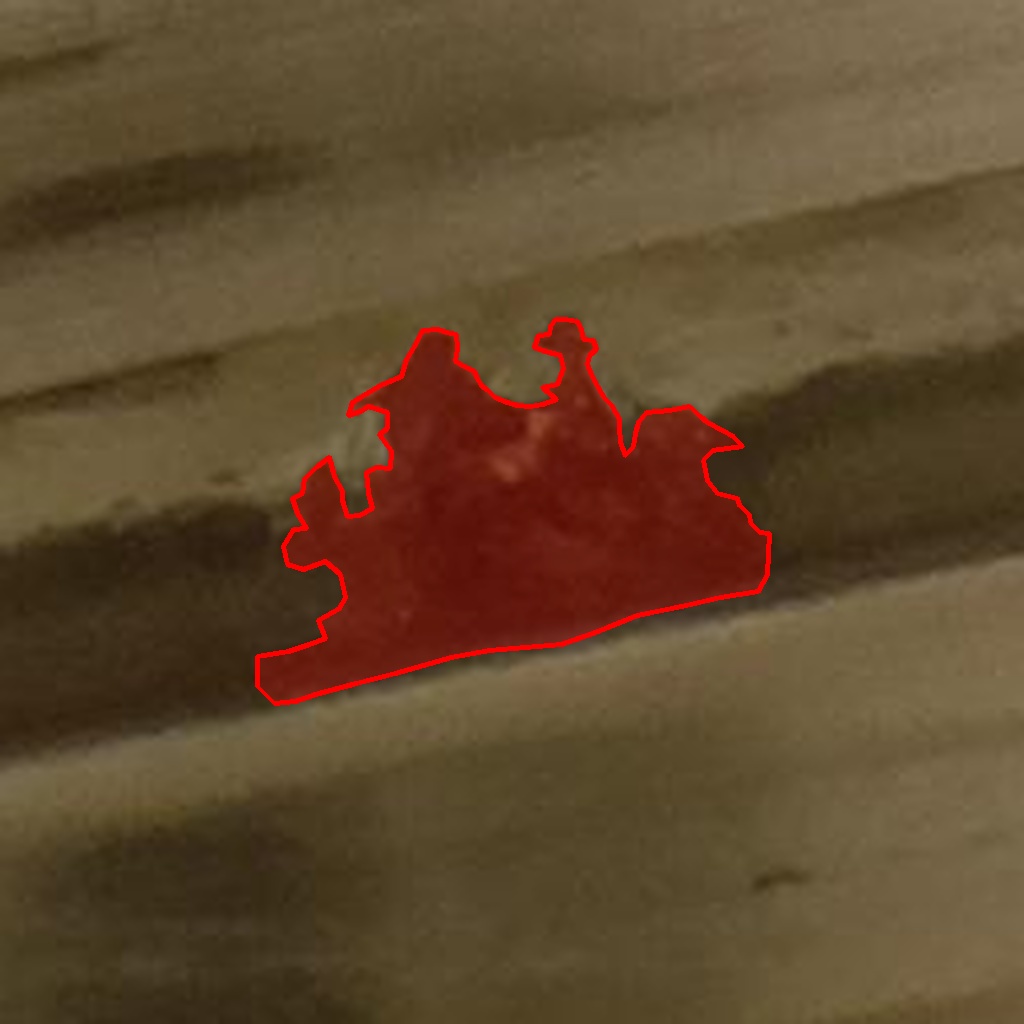}
    \caption{Example of a cropped defect patch with the automatically derived SAM segmentation mask overlaid.}
    \label{fig:defect_label}
\end{figure}
We extract defect patches from training images using segmentation masks for prompt generation (Fig.~\ref{fig:defect_label}), 
LoRA~\cite{hu2021loralowrankadaptationlarge} fine-tuning, and quality evaluation. 
Each instance is cropped with surrounding context and resized to $1024{\times}1024$\,px; 
when multiple defects overlap, only the largest is retained. 
This yields 221 patches. 
We provide a more detailed analysis of defect size and shape 
characteristics in the supplementary material (Sec.~A.1).

\noindent\textbf{MSD.}~\cite{zhang2022fdsnet} contains 1,200 images with three defect
types (oil, scratch, stain); for cross-dataset evaluation we use only the scratch subset
to assess pipeline transferability to a different defect type and domain
(Fig.~17 in the supplementary material). We apply the same split ratios as
BSData~\cite{schlagenhauf2021industrial}, yielding 260 training, 60 validation, and 80
test images with 890 scratch instances in the training set. For inpainting we use 20
randomly selected defect-free images, and for
LoRA~\cite{hu2021loralowrankadaptationlarge} training we use 100 randomly sampled
patches preprocessed as described for BSData~\cite{schlagenhauf2021industrial}.

\subsection{Synthetic Data Quality Analysis}\label{sec:synthdataqual}
This section describes how we select the optimal
LoRA~\cite{hu2021loralowrankadaptationlarge} variant and analyze prompt
construction and sample filtering for
BSData~\cite{schlagenhauf2021industrial}, using
DreamSim~\cite{fu2023dreamsim} and CLIPScore~\cite{hessel2021clipscore}
as quality metrics throughout. All LoRA variants follow the training
procedure described in Sec.~\ref{subsec:arch}.\\
\noindent\textbf{Qualitative comparison.}
During training, we generate intermediate samples every 400 steps with
the final prompt to monitor convergence and detect failure modes early.
We provide these samples in the supplementary (Fig.~14 in the supplementary material).\\
\noindent\textbf{Quantitative comparison.}
We generate synthetic patches with each LoRA variant and evaluate them
using DreamSim~\cite{fu2023dreamsim} (perceptual similarity to real
patches) and CLIPScore~\cite{hessel2021clipscore} (prompt alignment).
For DreamSim we report the minimum distance to the nearest real patch
(\textit{min\_dist}) and the mean distance to the $k{=}3$ nearest
neighbours (\textit{mean\_k\_dist}). The size-binned strategy produces
20 samples per bucket (60 total); the random strategy is evaluated on
an equal number under identical settings.
Tab.~\ref{tab:lora_choice} \textit{(a)} summarizes the results. The
LoRA trained on large defects achieves the best overall performance,
yielding the highest mean and median CLIPScore as well as the lowest
DreamSim distances for both metrics. Best value per metric column is
\textbf{\colorbox{yellow!50}{yellow}}.

\begin{figure}[!htbp]
  \centering
  \includegraphics[width=\textwidth]{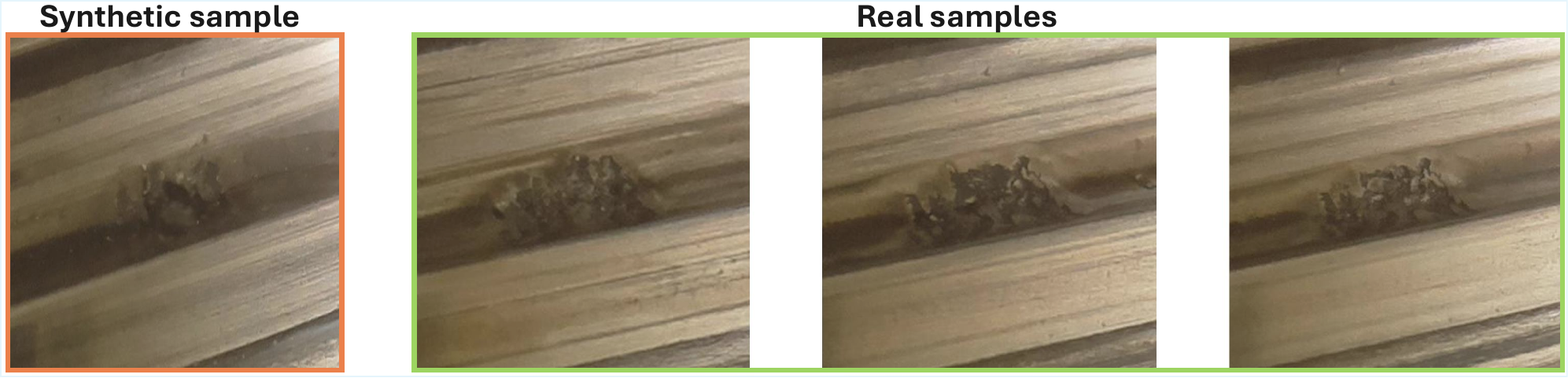}
    \caption{A synthetic defect sample generated by the top-performing 
    LoRA~\cite{hu2021loralowrankadaptationlarge} model, selected based 
    on CLIPScore \textit{(left)}, alongside its three nearest neighbors from the 
    training set, retrieved using DreamSim~\cite{fu2023dreamsim} 
    perceptual distance \textit{(right)}. This comparison validates that the 
    generated sample exhibits visual diversity and is not merely 
    replicating training data.}
  \label{fig:loradist}
\end{figure}

\noindent\textbf{LoRA Selection.}
Tab.~\ref{tab:lora_choice} \textit{(b)} compares the random
LoRA~\cite{hu2021loralowrankadaptationlarge} with the aggregated
size-binned strategy over $n{=}60$ samples each, showing that the
random variant achieves higher mean and median
CLIPScore~\cite{hessel2021clipscore} while the size-binned strategy
yields only marginally lower DreamSim~\cite{fu2023dreamsim} distances.
We therefore select the random LoRA for downstream synthesis, as it
offers the strongest prompt consistency, remains competitive in
perceptual similarity, and requires no size-specific model selection at
generation time. Fig.~\ref{fig:loradist} shows a representative
synthetic sample produced with the selected LoRA alongside its nearest
real training patches according to DreamSim distance.

\begin{table}[!htbp]
\scriptsize
\centering
\resizebox{\textwidth}{!}{%
\begin{tabular}{l|c|c|c|c|c|c}
\toprule
\textbf{Bucket/Strategy} & \textbf{$n$} & \textbf{CLIPScore mean $\pm$ std $\uparrow$} & \textbf{CLIPScore median $\uparrow$} & \textbf{min\_dist mean $\downarrow$} & \textbf{mean\_k\_dist mean $\downarrow$} & \textbf{mean\_k\_dist med. $\downarrow$} \\
\midrule
\multicolumn{7}{l}{\textit{(a) Size-binned strategy ($n{=}20$ per bucket)}} \\
\midrule
small & 20 & 26.0971 $\pm$ 1.0242 & 26.2082 & 0.2193 & 0.2275 & 0.2242 \\
medium & 20 & 26.8142 $\pm$ 1.6936 & 27.1947 & 0.2204 & 0.2293 & \cellcolor{yellow!50}\textbf{0.2128} \\
large & 20 & \cellcolor{yellow!50}\textbf{27.9290 $\pm$ 1.4032} & \cellcolor{yellow!50}\textbf{27.7938} & \cellcolor{yellow!50}\textbf{0.2079} & \cellcolor{yellow!50}\textbf{0.2158} & 0.2167 \\
\midrule
\multicolumn{7}{l}{\textit{(b) Aggregated comparison ($n{=}60$ per strategy)}} \\
\midrule
random & 60 & \cellcolor{yellow!50}\textbf{27.6974 $\pm$ 1.3535} & \cellcolor{yellow!50}\textbf{27.9839} & 0.2186 & 0.2248 & \cellcolor{yellow!50}\textbf{0.2169} \\
size & 60 & 26.9468 $\pm$ 1.5727 & 26.8120 & \cellcolor{yellow!50}\textbf{0.2159} & \cellcolor{yellow!50}\textbf{0.2242} & 0.2176 \\
\bottomrule
\end{tabular}%
}
\caption{BSData~\cite{schlagenhauf2021industrial} LoRA~\cite{hu2021loralowrankadaptationlarge} comparison.
\textit{(a)}~Size-binned strategy with per-bucket results; \textit{(b)}~aggregated comparison between random and size-binned strategies.}
\label{tab:lora_choice}
\end{table}

\noindent\textbf{Prompt Construction.}
Prompts are derived through automatic tag extraction using
Qwen2-VL~\cite{wang2024qwen2vl,yang2025qwen3technicalreport}. The
model processes batches of four defect patches and returns up to 15
tags describing material properties, defect morphology, texture, and
recording conditions. Tags are aggregated across all batches, ranked by
frequency, and lightly pruned to suppress overly generic terms. The
final conditioning string (Fig.~\ref{fig:prompt_qwen_summary}) is
assembled from the most informative descriptors, ordered by scene,
defect type, texture, and recording conditions. Full prompt details are
provided in the supplementary material
(Sec.~C in the supplementary material).

\noindent\textbf{Filtering.}
For BSData~\cite{schlagenhauf2021industrial}, we generate 1{,}000
candidate defect patches via mask-guided inpainting with
Flux.1-dev~\cite{flux2024,labs2025flux1kontextflowmatching} and the
selected LoRA~\cite{hu2021loralowrankadaptationlarge}. We then filter
this pool using DreamSim~\cite{fu2023dreamsim} and
CLIPScore~\cite{hessel2021clipscore} to balance perceptual similarity
and prompt fidelity. A detailed analysis of the score distributions and
their relationship to defect characteristics is provided in the
supplementary material~(Sec.~C in the supplementary material).

\FloatBarrier

\begin{figure}[!htbp]
\centering
\fbox{%
  \begin{minipage}{0.96\textwidth}
  \small
  \textbf{Resulting prompt:}\\[2pt]
  \textit{pitting defect on galvanized steel, irregular pitted surface, rough grainy texture, dark pits, subtle metallic sheen, close-up industrial inspection photo, shallow depth of field, low contrast, dim diffuse lighting, shadowed edges, horizontal striations}
  \end{minipage}%
}
\caption{BSData~\cite{schlagenhauf2021industrial} prompt derived from frequent Qwen~\cite{wang2024qwen2vl} 
tags and light manual pruning. The prompt emphasizes material, morphology, texture, 
and recording conditions relevant to pitting defects.}
\label{fig:prompt_qwen_summary}
\end{figure}

\subsection{Detection Performance}\label{subsec:detectperf}
We evaluate the effect of synthetic data on downstream defect detection
by training each detector under the configurations described in
Sec.~\ref{sec:trainingconfig} and testing on the fixed
BSData~\cite{schlagenhauf2021industrial} test split. All results are
reported as mean $\pm$ standard deviation over three runs. YOLOX~\cite{ge2021yolox}
results are provided in the supplementary material~(Sec.~C.1 in the supplementary material).
Best value per metric column is \textbf{\colorbox{yellow!50}{yellow}}, 2nd best \colorbox{gray!25}{gray} while \colorbox{blue!15}{blue} indicates the baseline.\\
\noindent\textbf{YOLOv26.}
Tab.~\ref{tab:test_yolo26_lwdetr} \textit{(left)} summarizes the
YOLOv26~\cite{sapkota2026yolo26keyarchitecturalenhancements} results.
The real-only baseline reaches $\text{AP}=0.652\pm0.006$.
Among mixed regimes, the 75/25 split performs best
($\text{AP}=0.655\pm0.026$), closely matching the baseline,
while increasing the synthetic fraction further degrades performance.
Synthetic-only training drops substantially to $\text{AP}=0.393\pm0.005$.
The strongest overall result is obtained in the union setting
(100\%\,R\,+\,100\%\,S) at 150 epochs ($\text{AP}=0.667\pm0.006$),
suggesting that synthetic data yields modest gains when the full real
set is retained. The same configuration at 300 epochs achieves the
best AP50 ($0.933\pm0.015$) and Recall ($0.878\pm0.045$).\\
\noindent\textbf{LW-DETR.}
Tab.~\ref{tab:test_yolo26_lwdetr} \textit{(right) }shows the
LW-DETR~\cite{chen2024lwdetrtransformerreplacementyolo} results.
Unlike YOLOv26, LW-DETR achieves its highest AP under the
real-only baseline ($\text{AP}=0.666\pm0.011$) and does not benefit
consistently from synthetic mixtures, with AP declining as the
synthetic fraction increases. Union settings partially recover
performance: 100\%\,R\,+\,100\%\,S at 90 epochs yields the best
AP50 ($0.951\pm0.014$), and 100\%\,R\,+\,200\%\,S at 90 epochs
achieves the best AR100 ($0.715\pm0.007$). Overall, LW-DETR is
more robust to synthetic additions but profits less from them,
indicating that the benefit of synthetic augmentation is
detector-dependent.

\begin{table*}[!htbp]
\scriptsize
\centering
\sisetup{separate-uncertainty=true}
\resizebox{\textwidth}{!}{%
\begin{tabular}{
    l
    c S[table-format=1.3(3)] S[table-format=1.3(3)] S[table-format=1.3(3)] S[table-format=1.3(3)]
    c S[table-format=1.3(3)] S[table-format=1.3(3)] S[table-format=1.3(3)] S[table-format=1.3(3)]
}
\toprule
& \multicolumn{5}{c}{\textbf{YOLOv26~\cite{sapkota2026yolo26keyarchitecturalenhancements}}}
& \multicolumn{5}{c}{\textbf{LW-DETR~\cite{chen2024lwdetrtransformerreplacementyolo}}} \\
\cmidrule(lr){2-6} \cmidrule(lr){7-11}
\textbf{Exp.(R/S)}
& \textbf{Ep.}
& \multicolumn{1}{c}{\textbf{AP~\up}}
& \multicolumn{1}{c}{\textbf{AP50~\up}}
& \multicolumn{1}{c}{\textbf{Prec.~\up}}
& \multicolumn{1}{c}{\textbf{Recall~\up}}
& \textbf{Ep.}
& \multicolumn{1}{c}{\textbf{AP~\up}}
& \multicolumn{1}{c}{\textbf{AP50~\up}}
& \multicolumn{1}{c}{\textbf{AP75~\up}}
& \multicolumn{1}{c}{\textbf{AR100~\up}} \\
\midrule
100/0
& 300 & \cellcolor{blue!15}\num{0.652 \pm 0.006} & \cellcolor{blue!15}\num{0.926 \pm 0.021} &\cellcolor{blue!15}\num{0.931 \pm 0.010} & \cellcolor{blue!15}\num{0.854 \pm 0.035}
& 90 & \cellcolor{blue!15}{ 0.666 $\pm$ 0.011} & \cellcolor{blue!15}\num{0.924 \pm 0.015} & \cellcolor{blue!15}{0.795 $\pm$ 0.019} & \cellcolor{blue!15}\num{0.709 \pm 0.010} \\
75/25
& 300 & \cellcolor{gray!25}\num{0.655 \pm 0.026} & \cellcolor{gray!25}\num{0.926 \pm 0.013} & \num{0.913 \pm 0.011} & \cellcolor{gray!25}\num{0.864 \pm 0.042}
& 90 & \num{0.656 \pm 0.006} & \num{0.909 \pm 0.024} & \cellcolor{gray!25}\num{0.792 \pm 0.013} & \num{0.697 \pm 0.004} \\
50/50
& 300 & \num{0.641 \pm 0.035} & \num{0.906 \pm 0.014} & \cellcolor{yellow!50}{\bfseries 0.954 $\pm$ 0.034} & \num{0.803 \pm 0.037}
& 90 & \num{0.642 \pm 0.005} & \num{0.912 \pm 0.004} & \num{0.776 \pm 0.030} & \num{0.689 \pm 0.007} \\
25/75
& 300 & \num{0.602 \pm 0.023} & \num{0.871 \pm 0.022} & \cellcolor{gray!25}\num{0.930 \pm 0.054} & \num{0.763 \pm 0.039}
& 90 & \num{0.611 \pm 0.015} & \num{0.880 \pm 0.019} & \num{0.715 \pm 0.016} & \num{0.651 \pm 0.018} \\
0/100
& 300 & \num{0.393 \pm 0.005} & \num{0.729 \pm 0.002} & \num{0.819 \pm 0.046} & \num{0.610 \pm 0.043}
& 90 & \num{0.456 \pm 0.003} & \num{0.775 \pm 0.021} & \num{0.538 \pm 0.022} & \num{0.568 \pm 0.006} \\
100/100
& 150 & \cellcolor{yellow!50}{\bfseries 0.667 $\pm$ 0.006} & \num{0.915 \pm 0.010} & \num{0.941 \pm 0.027} & \num{0.843 \pm 0.020}
& 45 & \num{0.654 \pm 0.005} & \num{0.933 \pm 0.011} &  \cellcolor{yellow!50}{\bfseries 0.797 $\pm$ 0.015} & \cellcolor{gray!25}\num{0.707 \pm 0.013} \\
100/100
& 300 & \num{0.659 \pm 0.018} & \cellcolor{yellow!50}{\bfseries 0.933 $\pm$ 0.015} & \num{0.945 \pm 0.007} & \cellcolor{yellow!50}{\bfseries 0.878 $\pm$ 0.045}
& 90 & \cellcolor{gray!25} \num{0.661 \pm 0.005} & \cellcolor{yellow!50}{\bfseries 0.951 $\pm$ 0.014} & \num{0.773 \pm 0.007} & \num{0.705 \pm 0.008} \\
100/200
& 100 & \num{0.620 \pm 0.009} & \num{0.923 \pm 0.019} & \num{0.917 \pm 0.049} & \num{0.832 \pm 0.051}
& 30 & \num{0.634 \pm 0.006} & \num{0.923 \pm 0.020} & \num{0.766 \pm 0.007} & \num{0.687 \pm 0.009} \\
100/200
& 300 & \num{0.646 \pm 0.052} & \num{0.913 \pm 0.031} & \num{0.930 \pm 0.062} & \num{0.827 \pm 0.014}
& 90 & \cellcolor{yellow!50}{\bfseries 0.663 $\pm$ 0.005} & \cellcolor{gray!25}{0.949 $\pm$ 0.011} & \num{0.785 \pm 0.015} & \cellcolor{yellow!50}{\bfseries 0.715 $\pm$ 0.007} \\
\bottomrule
\end{tabular}%
}
\caption{BSData~\cite{schlagenhauf2021industrial} test set results for YOLOv26~\cite{sapkota2026yolo26keyarchitecturalenhancements} and LW-DETR~\cite{chen2024lwdetrtransformerreplacementyolo} (mean $\pm$ std over 3 runs).
\emph{R} and \emph{S} denote the percentage of real and synthetic training data relative to the full BSData training split.}
\label{tab:test_yolo26_lwdetr}
\end{table*}

\subsection{Cross-Dataset Transfer and Qualitative Ablations}\label{sec:ablations}
We next assess how much of the end-to-end pipeline transfers to the
MSD~\cite{zhang2022fdsnet} scratch subset and qualitatively analyze the
impact of LoRA~\cite{hu2021loralowrankadaptationlarge} on defect appearance.
Prompt construction, mask generation, inpainting, filtering, and detector
training follow the BSData~\cite{schlagenhauf2021industrial} setup, while
label derivation is simplified because scratch annotations remain unstable
under SAM~3~\cite{carion2025sam3segmentconcepts}.
For MSD~\cite{zhang2022fdsnet}, we use the synthetic inpainting mask directly as the segmentation annotation— bypassing SAM3\cite{carion2025sam3segmentconcepts}, which does not reliably delineate scratch regions—accepting minor label noise when the inpainted scratch occupies only part of the masked region.
For MSD~\cite{zhang2022fdsnet}, we train a single LoRA~\cite{hu2021loralowrankadaptationlarge} for 2{,}000 steps on 100 randomly sampled scratch patches and filter the resulting 1{,}000 candidate images to 520 samples using DreamSim~\cite{fu2023dreamsim} and CLIPScore~\cite{hessel2021clipscore}.\\
\noindent\textbf{Prompt Construction.}
Prompt construction follows the BSData~\cite{schlagenhauf2021industrial}
protocol: Qwen2-VL~\cite{wang2024qwen2vl,yang2025qwen3technicalreport}
processes scratch patches in batches of four and predicts up to 15 tags
per batch. Tags are aggregated by frequency and lightly pruned before final prompt assembly
(Tab.~5 in the supplementary material), and the final prompt combines
display-surface attributes with scratch-specific morphology and imaging
cues (Fig.~19 in the supplementary material). From the training
segmentations we generate 1{,}000 synthetic masks using the same random
scaling and translation scheme as for BSData, but with placement priors
spanning the full display area.

\noindent\textbf{Detection results.}
Tab.~\ref{tab:cross_yolo26} summarizes the YOLOv26 results on the MSD
scratch subset. The real-only baseline achieves a strong
$\text{AP}=0.953$, reflecting that MSD constitutes an easier
benchmark than BSData. Mixed settings degrade performance
monotonically as the synthetic fraction increases, with the 75/25
split being the closest to the baseline ($\text{AP}=0.936$,
$\text{AP50}=0.990$, $\text{Recall}=0.975$), while synthetic-only
training drops substantially to $\text{AP}=0.500$. Among union
configurations, 100\%\,R\,+\,200\%\,S at 300 epochs partially
recovers performance ($\text{AP}=0.934$) and achieves the best
Precision across all non-baseline settings ($0.984$). Overall, the
transferred pipeline remains usable after dataset-specific annotation
adaptation, but does not improve over the strong real-data baseline,
suggesting that synthetic augmentation is less impactful when
real data is already sufficient and class appearance is
less ambiguous.
\begin{table}[!htbp]
\scriptsize
\centering
\begin{tabularx}{\textwidth}{
    >{\raggedright\arraybackslash}X
    S[table-format=3.0, table-column-width=1.5cm]
    S[table-format=1.3, table-column-width=2.2cm]
    S[table-format=1.3, table-column-width=2.2cm]
    S[table-format=1.3, table-column-width=2.2cm]
    S[table-format=1.3, table-column-width=2.2cm]
}
\toprule
\textbf{Exp.(R/S)}
& \multicolumn{1}{c}{\textbf{Epochs}}
& \multicolumn{1}{c}{\textbf{AP~\up}}
& \multicolumn{1}{c}{\textbf{AP50~\up}}
& \multicolumn{1}{c}{\textbf{Precision~\up}}
& \multicolumn{1}{c}{\textbf{Recall~\up}} \\
\midrule
100/0 & 300 & \cellcolor{blue!15} {0.953} & \cellcolor{blue!15} {0.995} & \cellcolor{blue!15} {0.990} & \cellcolor{blue!15} {0.982} \\
75/25 & 300 & \cellcolor{yellow!50}{\bfseries0.936} & \cellcolor{yellow!50} {\bfseries0.990} & \cellcolor{gray!25} 0.978 & \cellcolor{yellow!50} {\bfseries0.975} \\
50/50 & 300 & 0.913 & \cellcolor{gray!25}0.984 & 0.972 & 0.939 \\
25/75 & 300 & 0.872 & 0.970 & 0.966 & 0.907 \\
0/100 & 300 & 0.500 & 0.770 & 0.827 & 0.652 \\
100/100 & 150 & 0.910 & 0.990 & 0.976 & 0.946 \\
100/100 & 300 & 0.908 & 0.982 & 0.971 & 0.949 \\
100/200 & 100 & 0.912 & 0.980 & 0.974 & 0.945 \\
100/200 & 300 & \cellcolor{gray!25}0.934 & 0.990 & \cellcolor{yellow!50}{\bfseries0.984} &  \cellcolor{gray!25}0.957 \\
\bottomrule
\end{tabularx}
\caption{YOLOv26~\cite{sapkota2026yolo26keyarchitecturalenhancements} results on the 
MSD~\cite{zhang2022fdsnet} test set. \emph{R} and \emph{S} denote real and synthetic training images, 
relative to the full MSD training split, in \%.}
\label{tab:cross_yolo26}
\end{table}

\begin{figure}[!ht]
    \centering
    \subcaptionbox{Score: 31.59}{\includegraphics[width=0.22\textwidth]{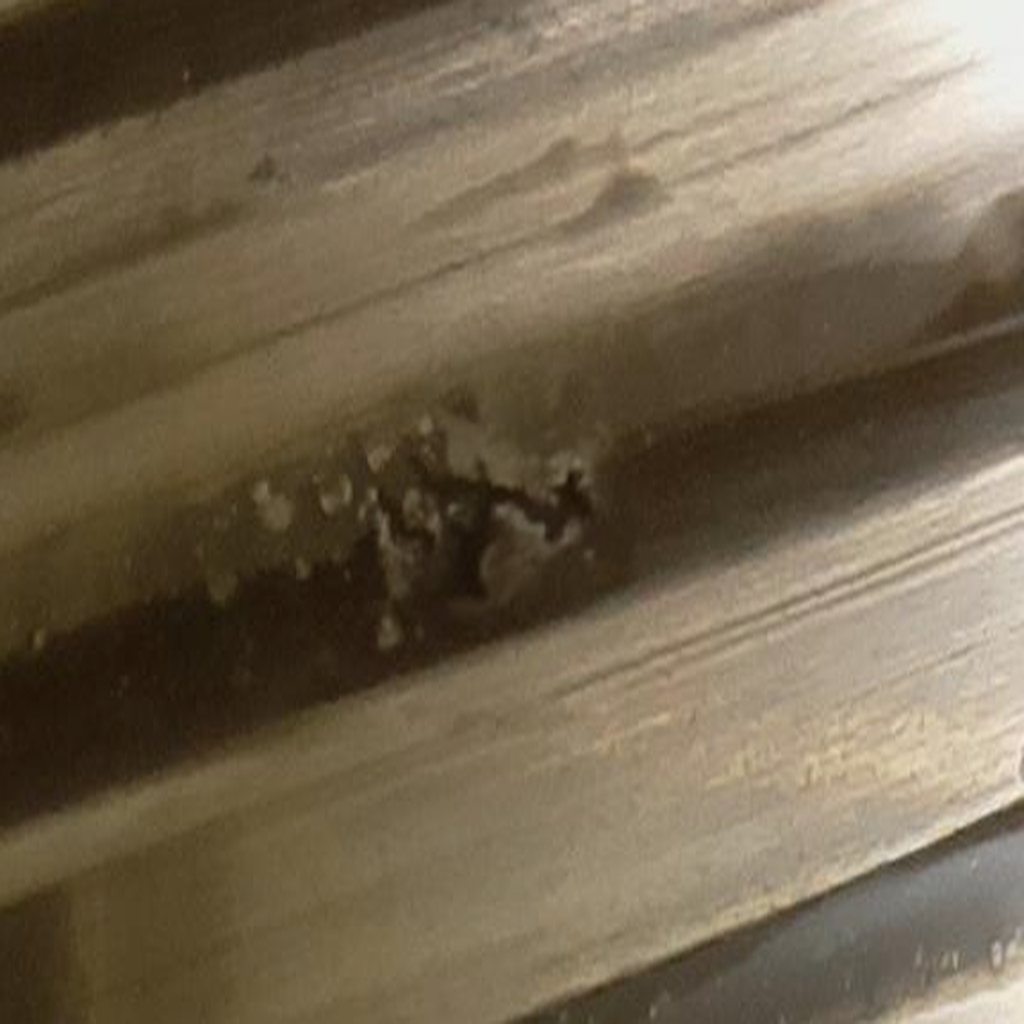}}
    \subcaptionbox{Score: 31.04}{\includegraphics[width=0.22\textwidth]{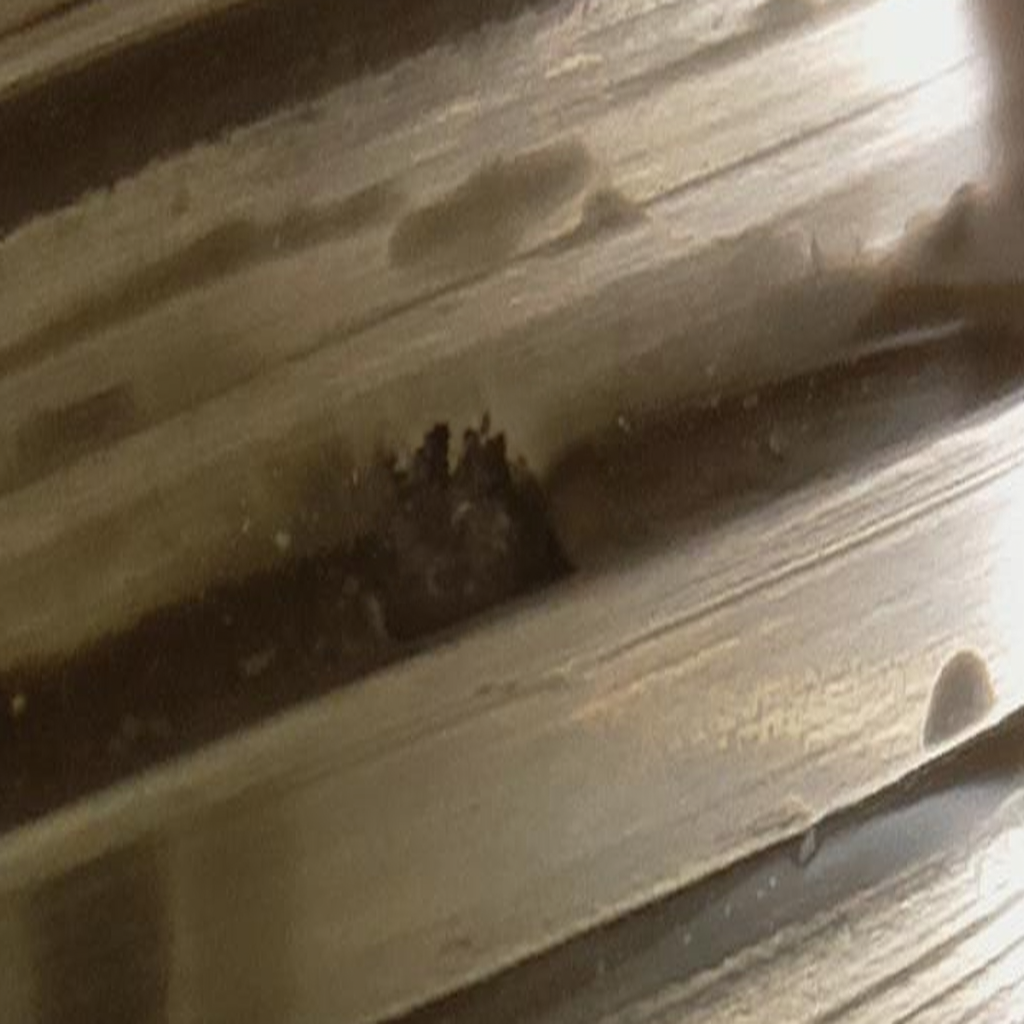}}
    \medskip
    \subcaptionbox{Score:17.17}{\includegraphics[width=0.22\textwidth]{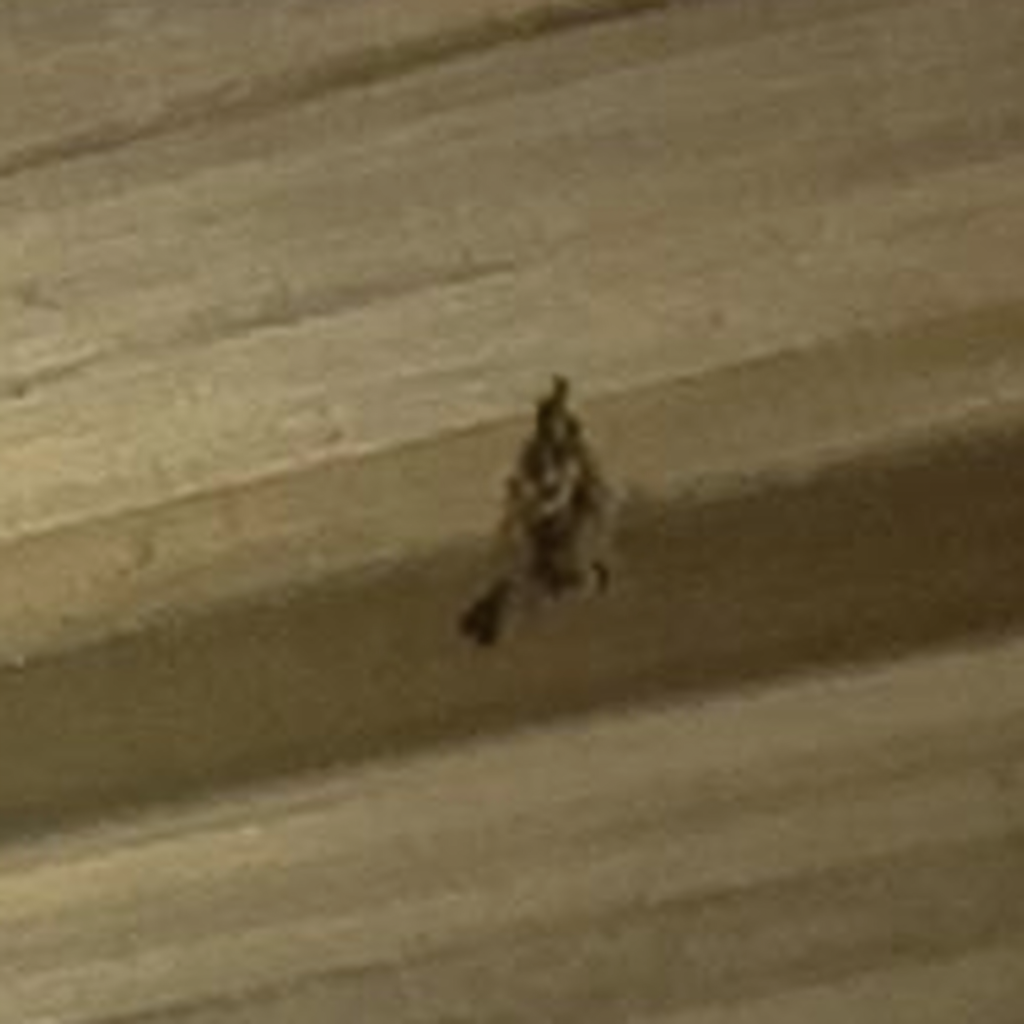}}
    \subcaptionbox{Score:17.40}{\includegraphics[width=0.22\textwidth]{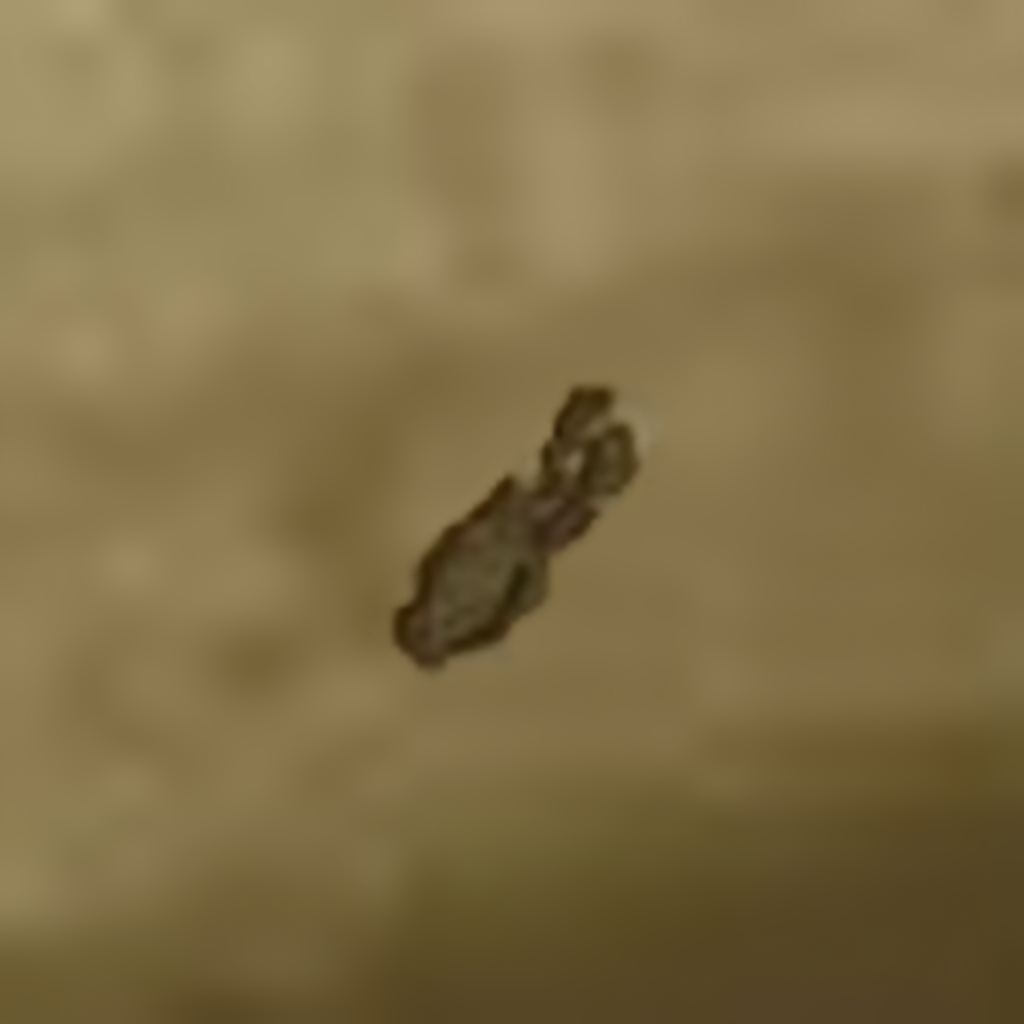}}
    \medskip
    \subcaptionbox{Sim: 0.129}{\includegraphics[width=0.22\textwidth]{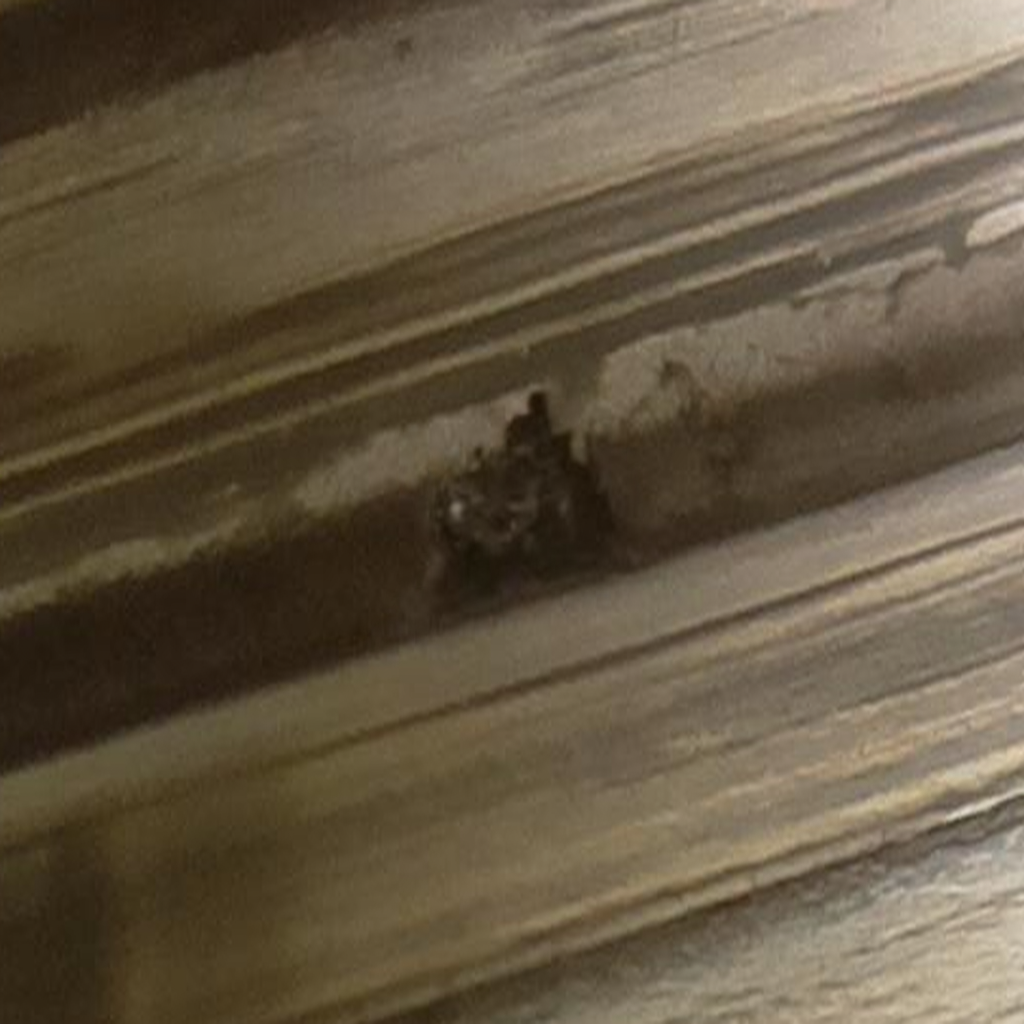}}
    \subcaptionbox{Sim: 0.128}{\includegraphics[width=0.22\textwidth]{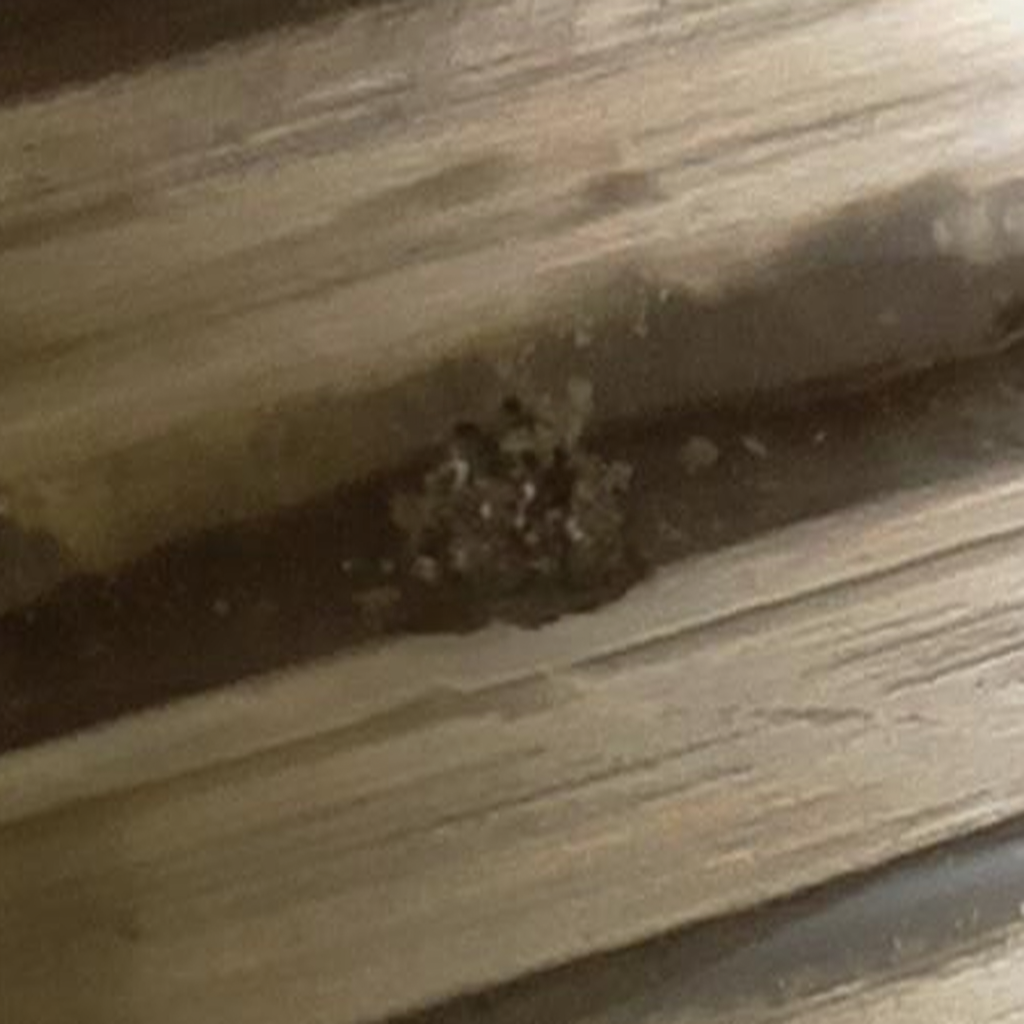}}
    \medskip
    \subcaptionbox{Sim: 0.459}{\includegraphics[width=0.22\textwidth]{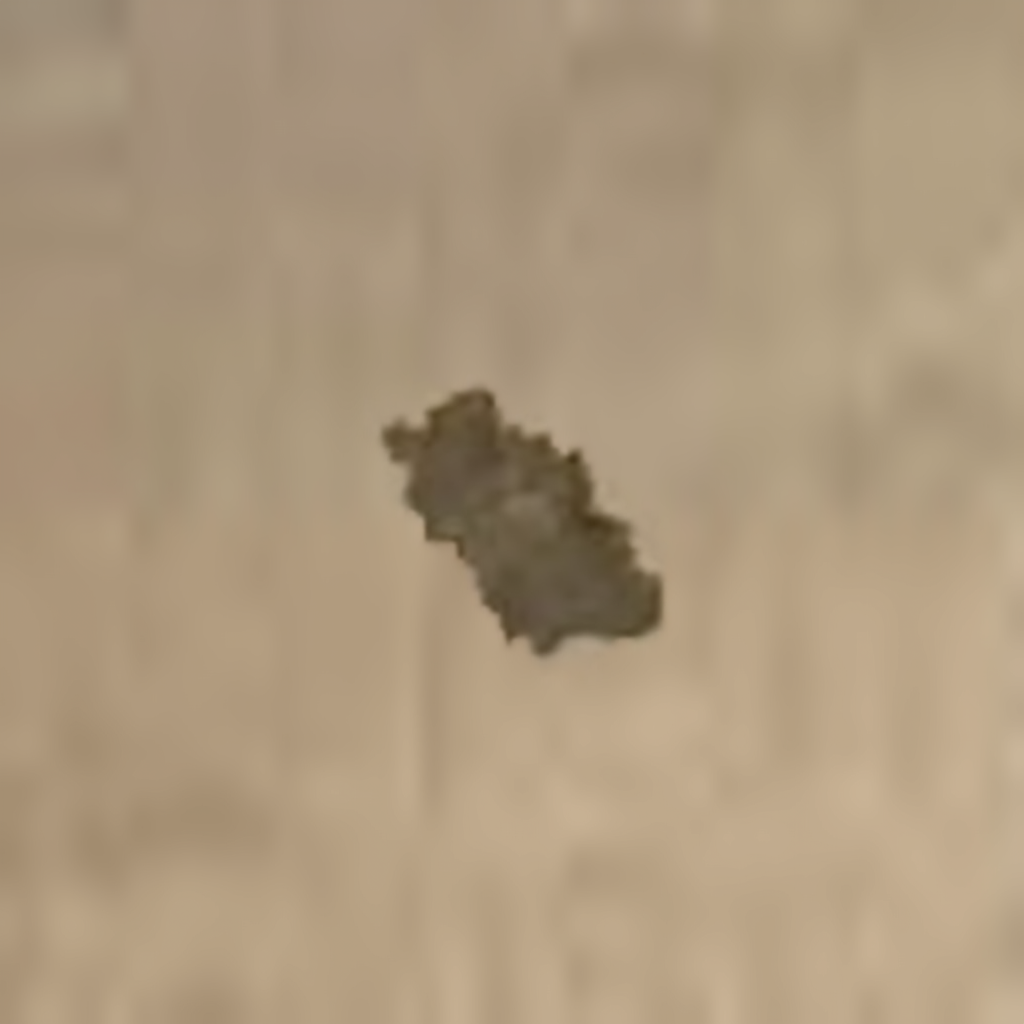}}
    \subcaptionbox{Sim: 0.447}{\includegraphics[width=0.22\textwidth]{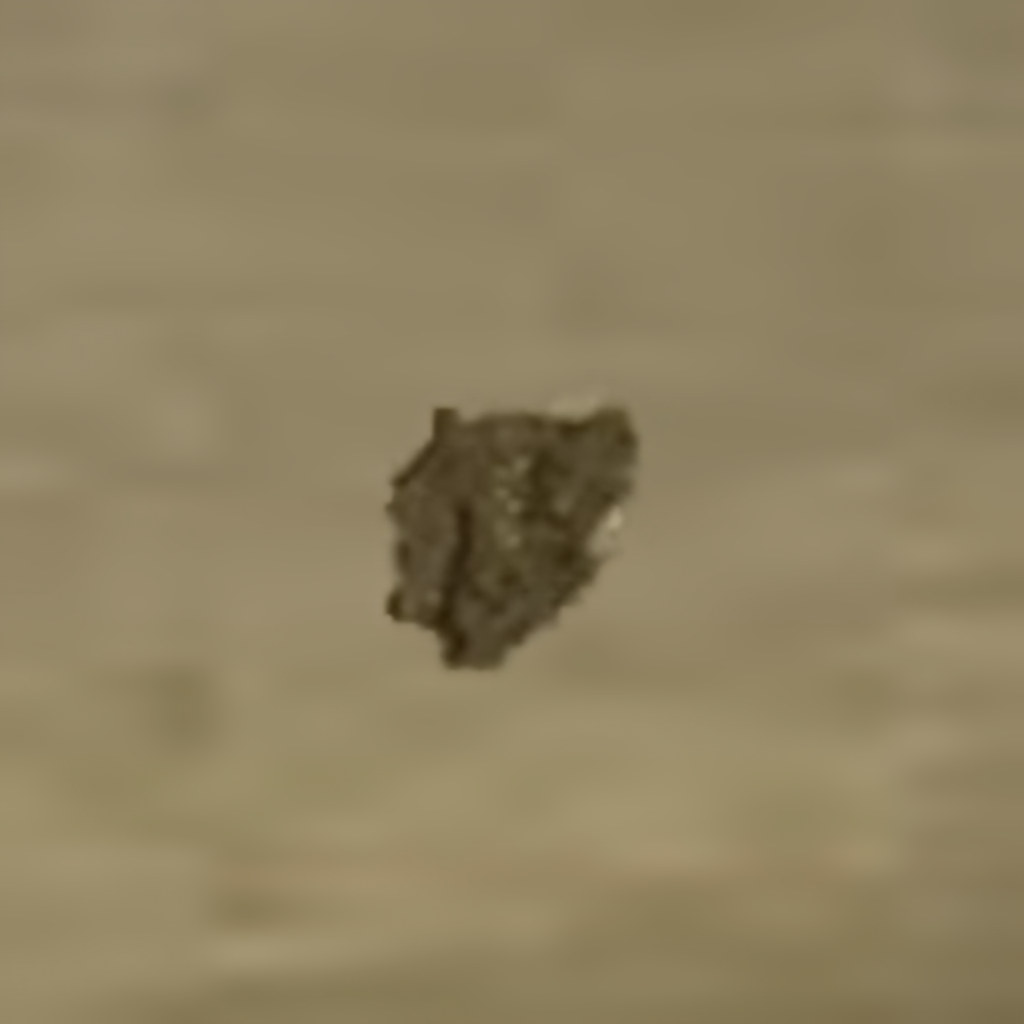}}
    
  \caption{Ranking extremes for synthetic patches on BSData~\cite{schlagenhauf2021industrial}.
    \textit{(a)},~\textit{(b)}~highest and \textit{(c)},~\textit{(d)}~lowest CLIPScore~\cite{hessel2021clipscore}~\up;
    \textit{(e)},~\textit{(f)}~closest and \textit{(g)},~\textit{(h)}~farthest DreamSim~\cite{fu2023dreamsim}~\down neighbors to real patches.
    Both metrics correlate, jointly identifying visually convincing samples with clear defect structure
    while suppressing blurred or weakly structured ones.}
\label{fig:ranking_extremes}
\end{figure}

\noindent\textbf{Synthesis without LoRA.}
Unconditional generations from Flux.1-dev~\cite{labs2025flux1kontextflowmatching,flux2024} without LoRA finetuning confirm that prompt optimization alone is insufficient to align the model with the target industrial domain (Fig.~15 in the supplementary material).\\
\noindent\textbf{Metric-based filtering analysis.}
To assess the effect of metric-based selection, we compare the highest- and lowest-ranked
generated patches under CLIPScore~\cite{hessel2021clipscore} and DreamSim~\cite{fu2023dreamsim}.
As shown in Fig.~\ref{fig:ranking_extremes}\,\textit{(a–d)}, high CLIPScore samples exhibit clearer
defect structure and more coherent local texture, while low-scoring samples tend to be blurred
or lack a distinct defect pattern — consistent with CLIPScore acting as a proxy for
prompt–image alignment. Fig.~\ref{fig:ranking_extremes} \textit{(e–h)} presents the corresponding
DreamSim extremes, measured as the mean $k$-NN distance to real defect patches. Samples with
low DreamSim distance closely resemble real training examples and frequently overlap with the
top CLIPScore selections; some remain visually subtle, which is plausible given the prevalence
of small, low-contrast defects in the training data. Conversely, high-distance DreamSim samples
are dominated by smooth or homogeneous textures, closely matching the weakest CLIPScore examples.
This strong correlation between both metrics confirms that they jointly serve as reliable
indicators of synthetic data quality, consistently favoring samples with clear,
domain-consistent defect structure and suppressing those with weak or implausible
defect appearance.

\subsection{Qualitative analysis}
\begin{figure}[!htbp]
    \centering
     \includegraphics[width=\textwidth]{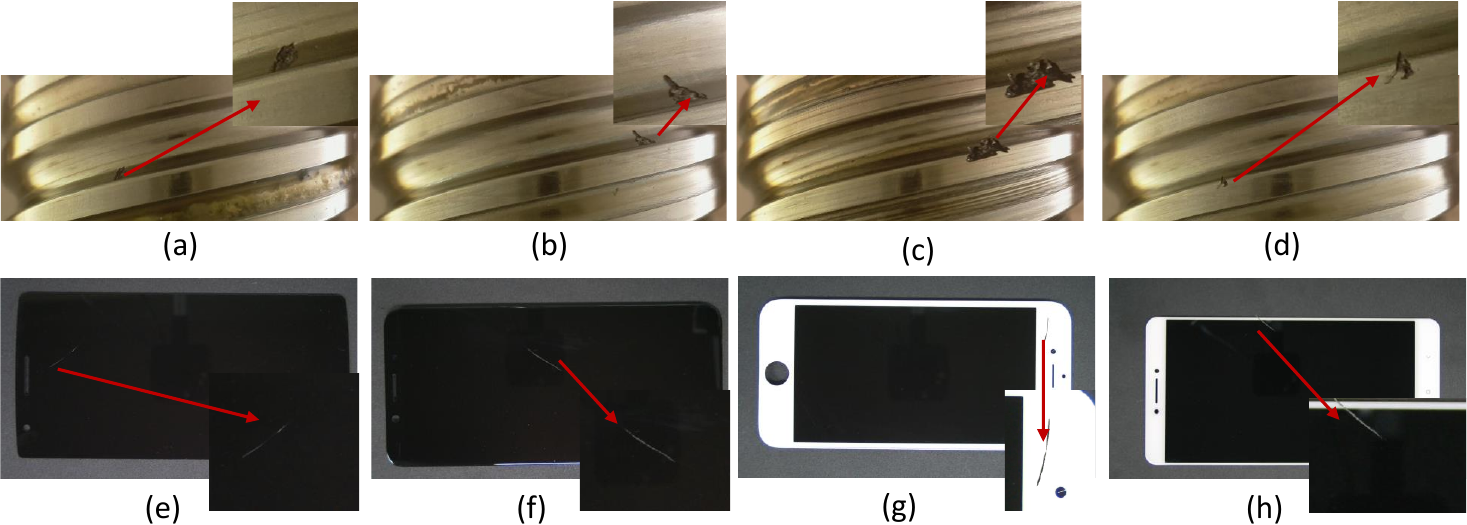}
\caption{Representative synthetic samples. \textbf{Top:} BSData~\cite{schlagenhauf2021industrial}; 
\textbf{Bottom:} MSD~\cite{zhang2022fdsnet}. Left \textit{(a,b,e,f)} show successful generations exhibiting 
plausible defect placement and realistic morphology. Right \textit{(c,d,g,h)} show typical failure cases, 
including boundary overlap, geometric distortion, and mask spillover artifacts.}
 \label{fig:bsd_examples}
\end{figure}

We conclude the qualitative analysis with representative successful and
unsuccessful synthetic samples, highlighting both realistic defect
characteristics and recurring failure modes that can degrade dataset
quality and annotation reliability.
For BSData~\cite{schlagenhauf2021industrial}, 
Fig.~\ref{fig:bsd_examples} \textit{(a-b)} successful pitting samples
exhibit plausible placement within the spindle groove and morphology
consistent with real defects—typically near-circular, locally
concentrated structures that may become slightly elongated at larger
scales. Failures occur when the synthetic mask overlaps component
boundaries or strong geometric edges, causing the inpainting process to
violate local surface geometry or produce structures no longer
resembling pitting defects Fig.~\ref{fig:bsd_examples} \textit{(c-d)}.
For MSD~\cite{zhang2022fdsnet}, successful samples display long, thin
scratches with subtle intensity variations consistent with the
underlying screen texture Fig.~\ref{fig:bsd_examples} \textit{(e-f)}, placed entirely within the display region.
Failures typically arise when the mask extends beyond the screen
area—e.g.\ due to varying screen dimensions across background
images—causing the generated scratch to partially fall outside the valid
inspection area Fig.~\ref{fig:bsd_examples} \textit{(g-h)}.

\section{Conclusion}\label{sec:conclusion}
We presented SynSur, a fully automated end-to-end pipeline for generating annotated synthetic defect images that addresses the labeled-data bottleneck in industrial visual inspection. Combining VLM-based prompting, metric-guided filtering, and SAM3~\cite{carion2025sam3segmentconcepts}–based label derivation, it requires no manual prompt engineering, annotation, or sample curation. On BSData~\cite{schlagenhauf2021industrial}, synthetic augmentation yields consistent AP improvements and substantially narrows the gap to full real-data performance in scarce-data regimes — the most practically relevant setting. LoRA proved essential for domain-consistent defect appearance, and the MSD~\cite{zhang2022fdsnet} transfer study confirms pipeline portability at modest engineering cost.\\
\noindent\textbf{Limitations and Future Work.} Synthetic-only training remains inferior to real-only training, and augmentation benefits are sensitive to mask quality, domain-specific tuning, and detector architecture. Further limitations include the single-defect-type focus and global prompting without controlled variation. Future work will pursue multi-class synthesis, learned placement priors, prompt diversification, adaptive mask generation, automated LoRA selection, and closed-loop refinement driven by detector feedback.
\appendix
\section{Supplemental Material}
We provide supplementary materials supporting the main findings of SynSur: An end-to-end generative pipeline for synthetic industrial surface defect generation and detection. It includes additional qualitative examples of synthetically generated surface defects, detailed architectural configurations of the generative pipeline, extended quantitative results across different defect categories, and further ablation studies not covered in the main text. These materials are intended to offer deeper insight into the pipeline's performance and reproducibility of the reported experiments.
\subsection{BSData dataset analysis.}\label{sec:appendix_bsd}
\begin{figure}[!htbp]
    \centering
    \includegraphics[width=0.4\textwidth]{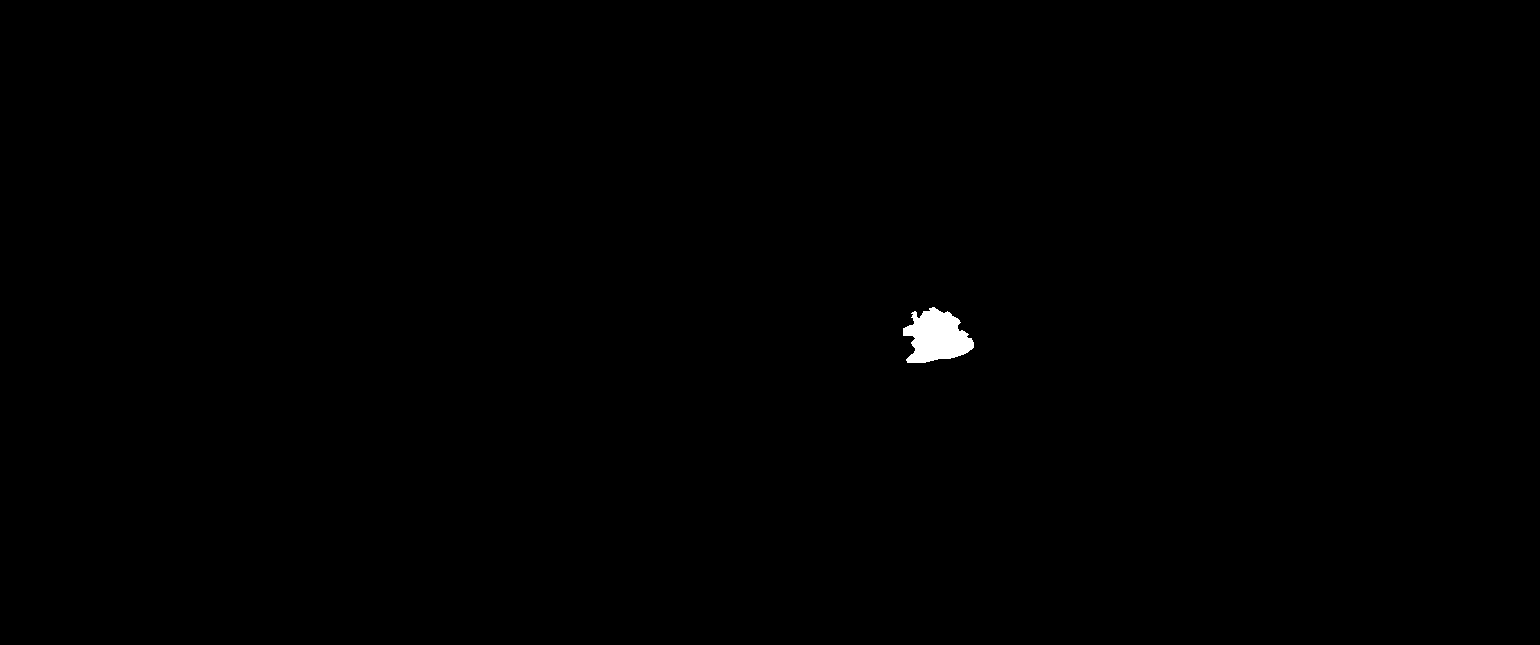}
    \hfill
    \includegraphics[width=0.4\textwidth]{images/A6149_1540x645_0220_n1_seed262.png}
    \caption{Examples of synthetic inpainting masks derived from real defect annotations. The masks preserve realistic defect morphology while allowing controlled spatial variation through scaling and shifting.}
    \label{fig:synth_masks}
\end{figure}

\begin{figure}[ht]
    \centering
    \begin{subfigure}[b]{0.49\textwidth}
        \centering
        \includegraphics[width=\linewidth]{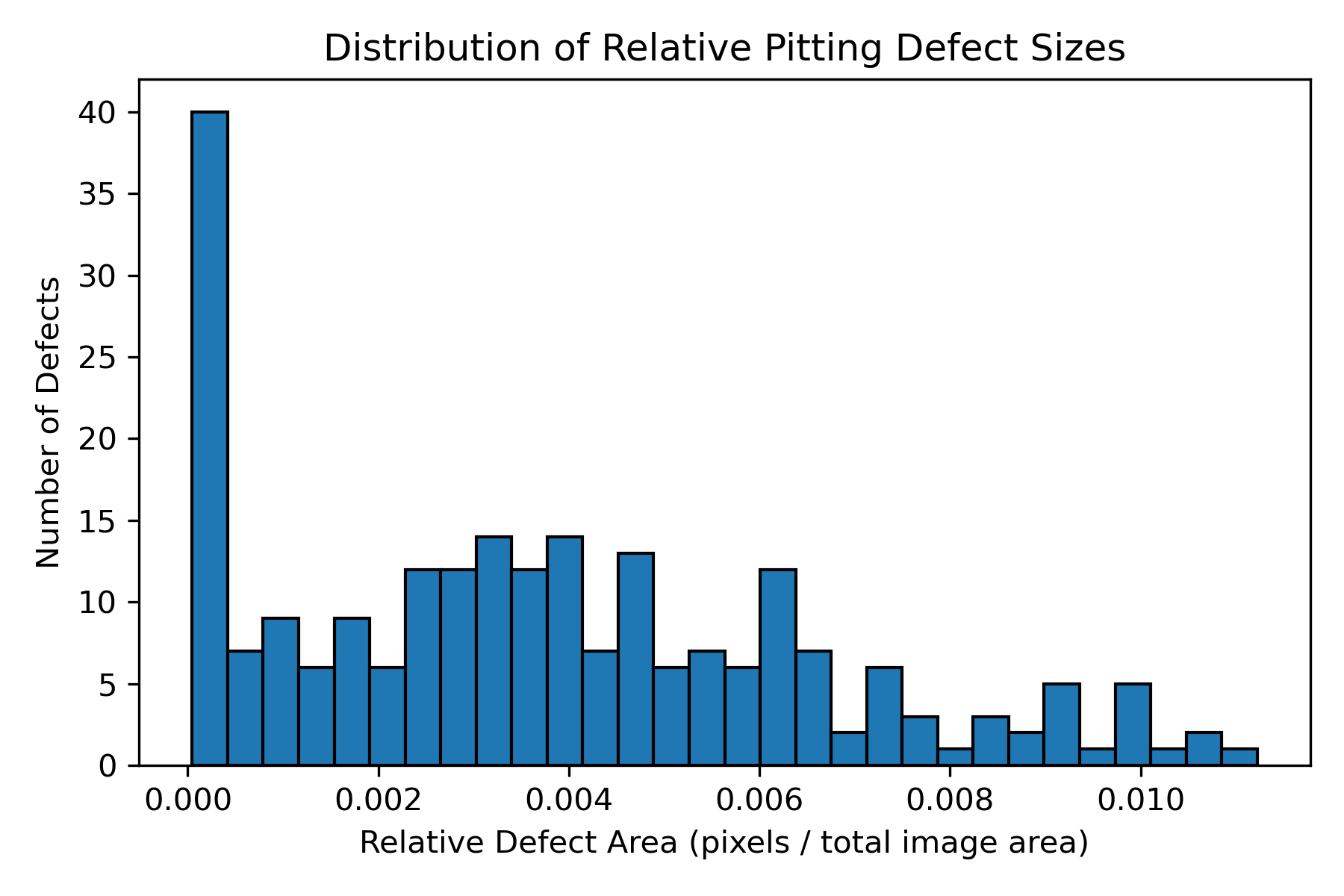}
        \caption{Relative defect area (instance mask area divided by image area).}
        \label{fig:relative_area}
    \end{subfigure}
    \hfill
    \begin{subfigure}[b]{0.49\textwidth}
        \centering
        \includegraphics[width=\linewidth]{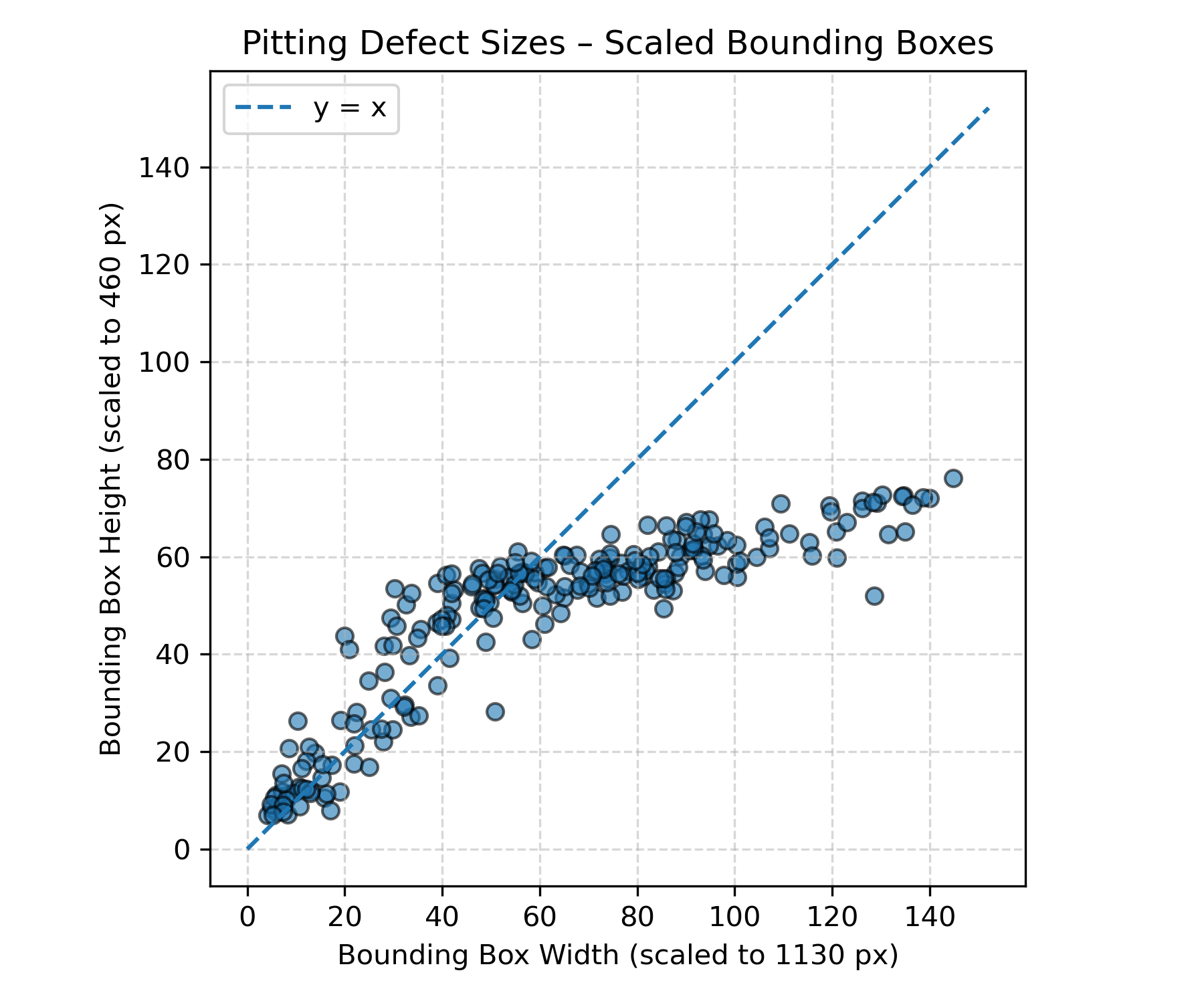}
        \caption{Bounding box aspect ratio vs.\ relative defect area.}
        \label{fig:bbox_scatter}
    \end{subfigure}
    \caption{Defect size and shape characteristics.}
    \label{fig:defect_sizes}
\end{figure}

\begin{figure}[ht]
    \centering
    \begin{subfigure}[b]{0.47\textwidth}
        \centering
        \includegraphics[width=\linewidth]{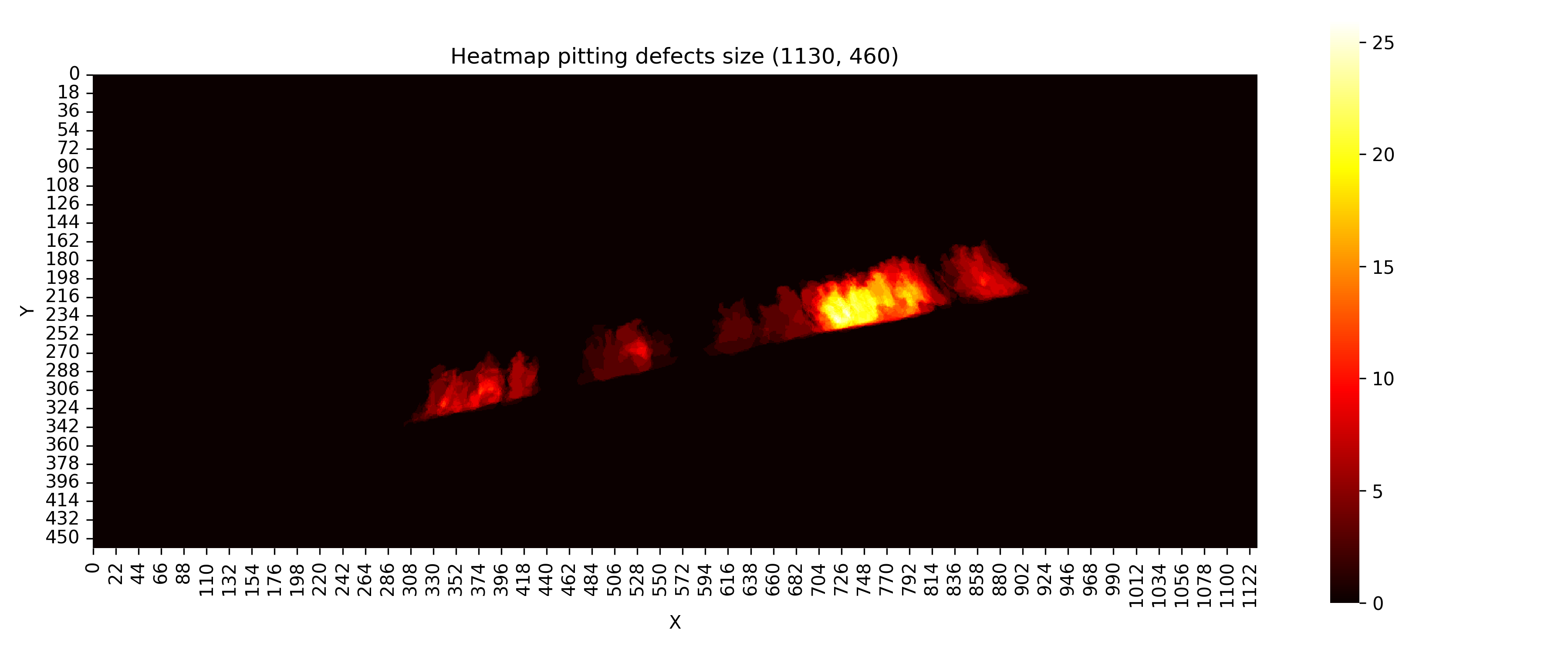}
        \caption{}
        \label{fig:heatmap_1130}
    \end{subfigure}
    \begin{subfigure}[b]{0.47\textwidth}
        \centering
        \includegraphics[width=\linewidth]{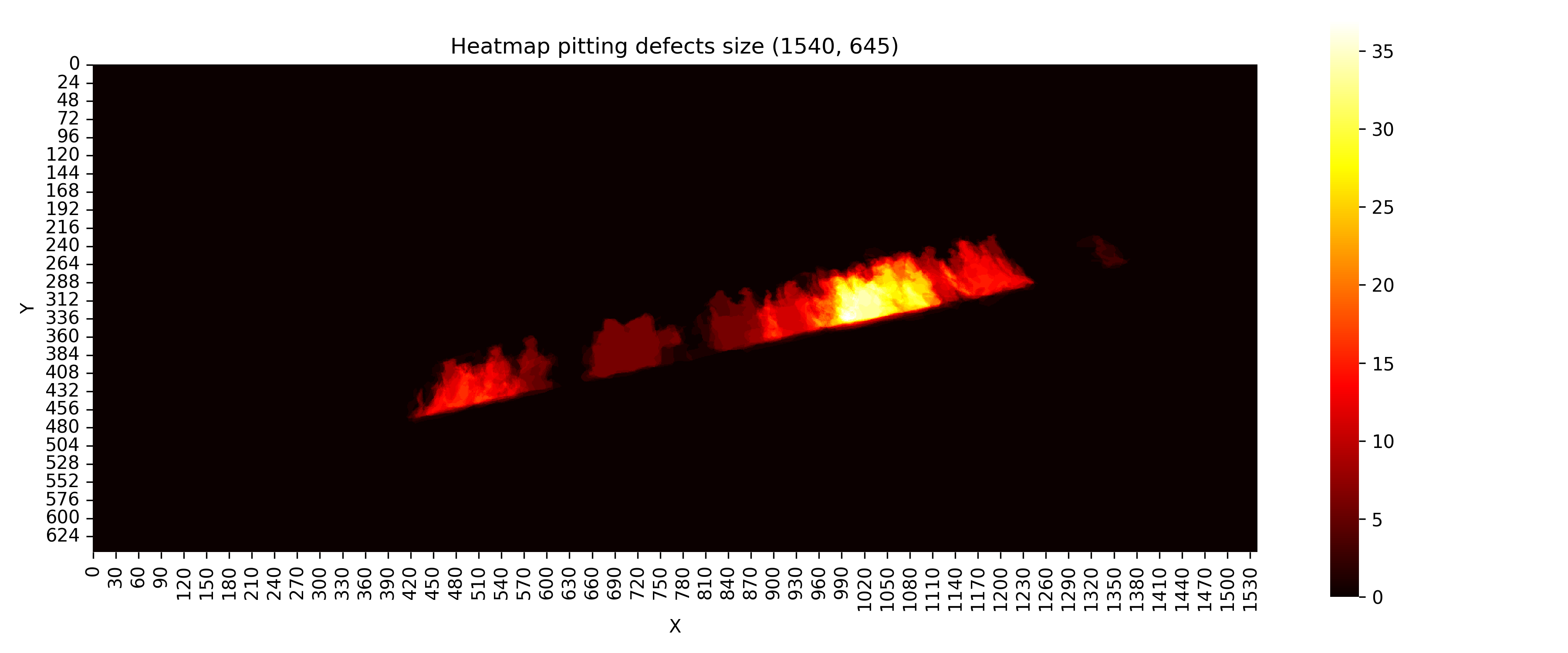}
        \caption{}
        \label{fig:heatmap_1540}
    \end{subfigure}
    
    \caption{Heatmaps of defect locations for the two retained image resolutions 
    (\emph{$1130\times460$} \textbf{(a)} and \emph{$1540\times645$} \textbf{(b)} respectively).  
    Defects are predominantly located in the upper left quadrant.}
    \label{fig:defect_positions}
\end{figure}

Synthetic mask examples are shown in Fig.~\ref{fig:synth_masks}.
Defects are concentrated in the central groove of the spindle (Fig.~\ref{fig:defect_positions})
and occupy only a small fraction of the image area (Fig.~\ref{fig:relative_area}).
Smaller defects appear compact and approximately circular,
whereas larger ones exhibit horizontal elongation (Fig.~\ref{fig:bbox_scatter}).

We filter the dataset to the two most frequent resolutions ($1130{\times}460$ and $1540{\times}645$ pixels) 
for consistent processing, resulting in 1,035 images: 710 defect-free and 325 defective, 
containing 357 annotated defect instances leading to a training data distribution of 65/15/20\%. 
Most defective images contain a single defect (299), 
while 20 contain two and 6 contain three instances. We analyze the spatial distribution of defects 
in the retained images in Fig.~\ref{fig:heatmap_1130} and~\ref{fig:heatmap_1540}, which shows a strong bias toward the upper left quadrant, 
likely due to the recording setup. This non-uniform distribution motivates our mask generation 
strategy in Fig.~\ref{fig:defect_positions} (\textbf{a-b}) to ensure realistic defect placement.

\section{Pipeline Efficiency.}
Industrial applicability requires that the pipeline runs without prohibitive overhead. All stages were executed on a single NVIDIA A100 40GB GPU with bf16 precision. LoRA fine-tuning for 2,000 steps on 100 patches at 1024×1024 px completes in approximately 2 hours (~3–4 seconds per step). Generating 1,000 candidate inpainting patches at 30 sampling steps each requires approximately 5.5 hours (~20 seconds per image). Metric-based filtering using DreamSim and CLIPScore over the full candidate pool takes approximately 20 minutes. SAM 3 segmentation and COCO annotation derivation for the 420 retained patches adds approximately 15 minutes. End-to-end, the pipeline delivers a fully annotated synthetic training set of 420 defect images in approximately 8 GPU-hours, with zero manual labeling effort beyond the seed annotations already required for LoRA training and mask derivation. 
\section{Synthetic Data Analysis}\label{sec:appendix_synthdata}

\begin{figure}[!htbp]
    \centering
    \begin{subfigure}[b]{0.49\textwidth}
        \centering
        \includegraphics[width=\linewidth]{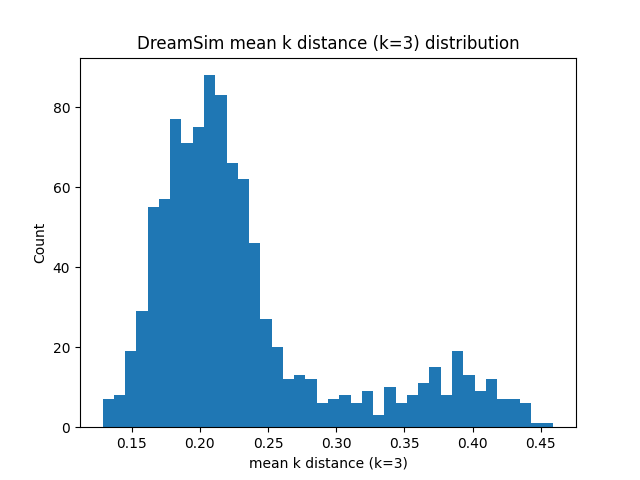}
        \caption{DreamSim.}\label{fig:dreamsim_dist}
    \end{subfigure}
    \hfill
    \begin{subfigure}[b]{0.49\textwidth}
        \centering
        \includegraphics[width=\linewidth]{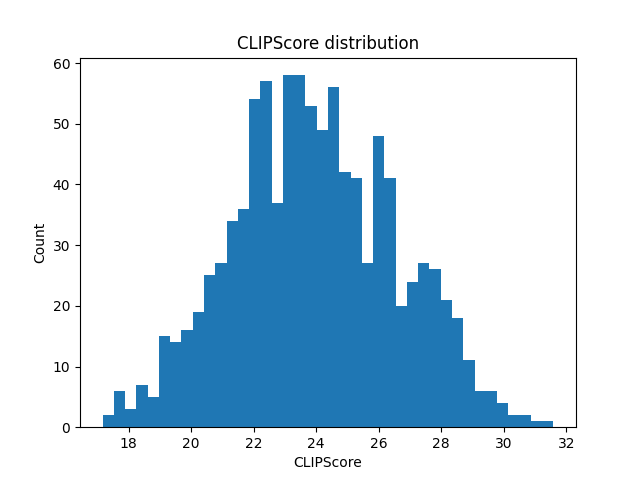}
        \caption{CLIPScore.}\label{fig:clipscore_dist}
    \end{subfigure}
    \caption{Score distributions for 1{,}000 generated BSData~\cite{schlagenhauf2021industrial}
candidate patches. DreamSim~\cite{fu2023dreamsim} exhibits a bimodal distribution,
whereas CLIPScore~\cite{hessel2021clipscore} is concentrated around a single mode.}
    \label{fig:dreamclip}
\end{figure}

\begin{figure}[!htbp]
    \centering
    \begin{subfigure}[b]{0.49\textwidth}
        \centering
        \includegraphics[width=\linewidth]{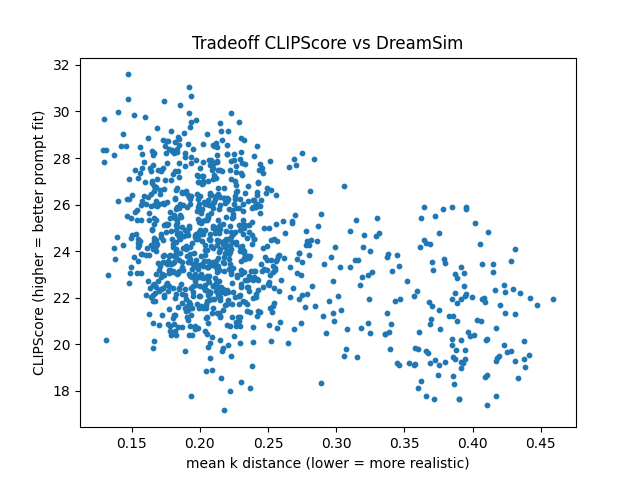}
        \caption{1{,}000 unfiltered samples.}\label{fig:unfiltered}
    \end{subfigure}
    \hfill
    \begin{subfigure}[b]{0.49\textwidth}
        \centering
        \includegraphics[width=\linewidth]{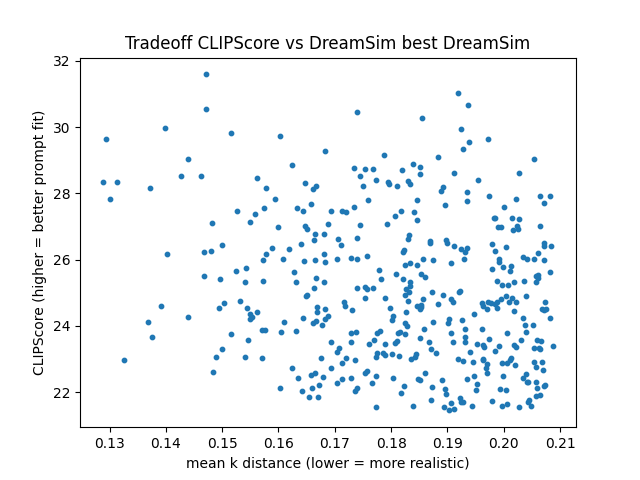}
        \caption{420 filtered samples.}\label{fig:filtered}
    \end{subfigure}
    \caption{Joint DreamSim~\cite{fu2023dreamsim} and CLIPScore~\cite{hessel2021clipscore} 
    ranking for BSData~\cite{schlagenhauf2021industrial} candidate patches before and after filtering. 
    The selected subset concentrates on the high-quality cluster.}
    \label{fig:tradeoff}
\end{figure}

\textbf{Prompt construction using automatically derived VLM tags.}
Prompts are constructed from tags automatically extracted by a
Qwen2-VL model~\cite{wang2024qwen2vl,yang2025qwen3technicalreport}.
Tab.~\ref{tab:qwen_tag_frequencies_top25} lists the most frequent extracted tags.
\begin{table}[!htbp]
\scriptsize
\centering
\sisetup{table-number-alignment=center}
\begin{tabularx}{\textwidth}{>{\raggedright\arraybackslash}X S[table-format=3.0] S[table-format=1.1]}
\toprule
\textbf{Tag} & \multicolumn{1}{c}{\textbf{Count}} & \multicolumn{1}{c}{\textbf{Share (\%)}} \\
\midrule
galvanized steel & 58 & 5.4 \\
irregular & 53 & 4.9 \\
close-up & 50 & 4.7 \\
rough & 43 & 4.0 \\
industrial & 43 & 4.0 \\
dark brown & 42 & 3.9 \\
pitting defect & 37 & 3.4 \\
dark & 36 & 3.4 \\
textured & 34 & 3.2 \\
horizontal & 34 & 3.2 \\
uneven & 32 & 3.0 \\
low contrast & 30 & 2.8 \\
surface & 27 & 2.5 \\
rough texture & 25 & 2.3 \\
high contrast & 24 & 2.2 \\
metallic & 23 & 2.1 \\
corroded & 18 & 1.7 \\
textured surface & 18 & 1.7 \\
pitting & 18 & 1.7 \\
industrial surface & 15 & 1.4 \\
dim lighting & 15 & 1.4 \\
surface wear & 14 & 1.3 \\
horizontal lines & 14 & 1.3 \\
low resolution & 12 & 1.1 \\
surface damage & 12 & 1.1 \\
\bottomrule
\end{tabularx}
\caption{Most frequent Qwen~\cite{wang2024qwen2vl} tags for the BSD~\cite{schlagenhauf2021industrial} pitting defect subset (top-25; total tag occurrences $N=3175$). 
Share is given in \%.}
\label{tab:qwen_tag_frequencies_top25}
\end{table}
Fig.~\ref{fig:dreamclip} shows the score distributions for all 1{,}000 candidates.
The CLIPScore~\cite{hessel2021clipscore} distribution (Fig.~\ref{fig:clipscore_dist})
is approximately unimodal, whereas DreamSim~\cite{fu2023dreamsim}
(Fig.~\ref{fig:dreamsim_dist}) exhibits two modes around 0.20 and 0.40.
This bimodality is also apparent in the joint score space
(Fig.~\ref{fig:unfiltered}), suggesting a natural split between higher-
and lower-quality generations. Applying a DreamSim threshold of 0.3
retains 839 samples. Removing the bottom 10\,\% by CLIPScore yields
755 samples, from which we select the 420 images with the highest
perceptual scores, corresponding to approximately twice the number
of real training defect patches (Fig.~\ref{fig:filtered}).

\begin{figure}[!htbp]
    \centering
    \begin{subfigure}[b]{0.49\textwidth}
        \centering
        \includegraphics[width=\linewidth]{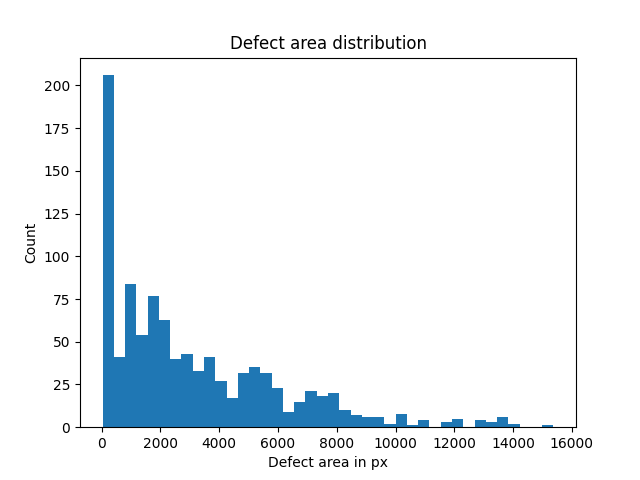}
        \caption{1{,}000 generated defect patches.}
        \label{fig:area_generated}
    \end{subfigure}
    \hfill
    \begin{subfigure}[b]{0.49\textwidth}
        \centering
        \includegraphics[width=\linewidth]{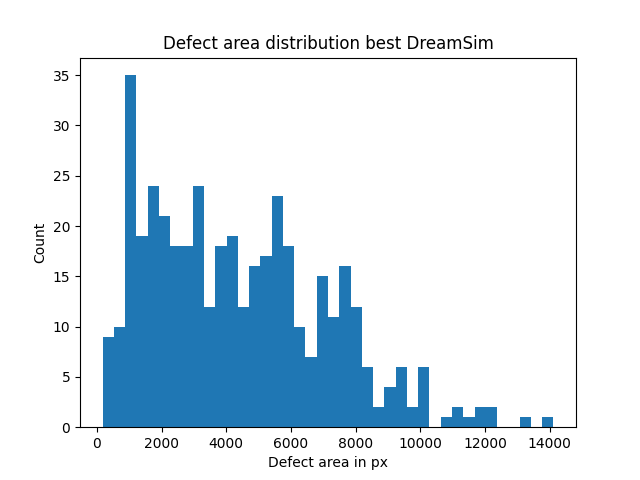}
        \caption{420 filtered patches.}\label{fig:area_filtered}
    \end{subfigure}
    \medskip
    \begin{subfigure}[b]{0.5\textwidth}
        \centering
        \includegraphics[width=\linewidth]{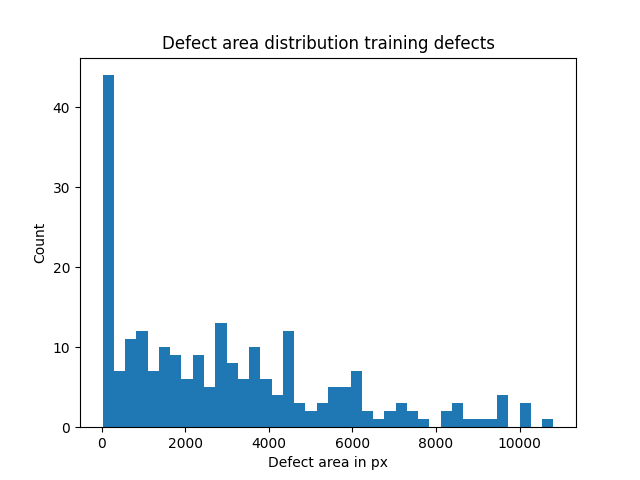}
        \caption{221 training defect patches.}\label{fig:area_train_data}
    \end{subfigure}
    \caption{Defect area distributions for BSData~\cite{schlagenhauf2021industrial} generated~\ref{fig:area_generated}, 
    filtered~\ref{fig:area_filtered}, and training patches~\ref{fig:area_train_data}. Filtering reduces the over representation of very small synthetic defects.}
    \label{fig:area}
\end{figure}

Fig.~\ref{fig:area} compares the defect-area distribution before and after filtering.
The original 1{,}000 candidates are dominated by small defects,
whereas the filtered subset is more uniformly distributed across defect areas.
The real training data remains skewed toward small defects but exhibits a smoother tail toward larger instances.\\\\
\begin{table}[!htbp]
\scriptsize
\centering
\sisetup{table-number-alignment=center}
\begin{tabularx}{\textwidth}{>{\raggedright\arraybackslash}X S[table-format=3.0] S[table-format=1.1]}
\toprule
\textbf{Tag} & \multicolumn{1}{c}{\textbf{Count}} & \multicolumn{1}{c}{\textbf{Share (\%)}} \\
\midrule
high contrast & 147 & 4.6 \\
sharp edges & 143 & 4.5 \\
glass & 130 & 4.1 \\
close-up view & 124 & 3.9 \\
metallic sheen & 124 & 3.9 \\
isolated defect & 121 & 3.8 \\
dark background & 111 & 3.5 \\
reflective surface & 89 & 2.8 \\
fine texture & 78 & 2.5 \\
uniform lighting & 77 & 2.4 \\
linear scratch & 73 & 2.3 \\
reflective & 68 & 2.1 \\
single defect & 67 & 2.1 \\
minimal background & 62 & 2.0 \\
diagonal orientation & 59 & 1.9 \\
metal surface & 55 & 1.7 \\
steel & 54 & 1.7 \\
scratches & 53 & 1.7 \\
matte finish & 50 & 1.6 \\
close-up & 50 & 1.6 \\
rust spots & 48 & 1.5 \\
clear focus & 46 & 1.4 \\
surface defect & 45 & 1.4 \\
blurred background & 45 & 1.4 \\
macro shot & 44 & 1.4 \\
\bottomrule
\end{tabularx}
\caption{Most frequent Qwen~\cite{wang2024qwen2vl} tags for the MSD~\cite{zhang2022fdsnet} scratch subset (top-25; total tag occurrences $N=3175$). 
Share is given in \%.}
\label{tab:qwen_tag_freq_msd_top25}
\end{table}

\begin{figure}[!htbp]
    \centering
    \subcaptionbox{0}{\includegraphics[width=0.15\textwidth]{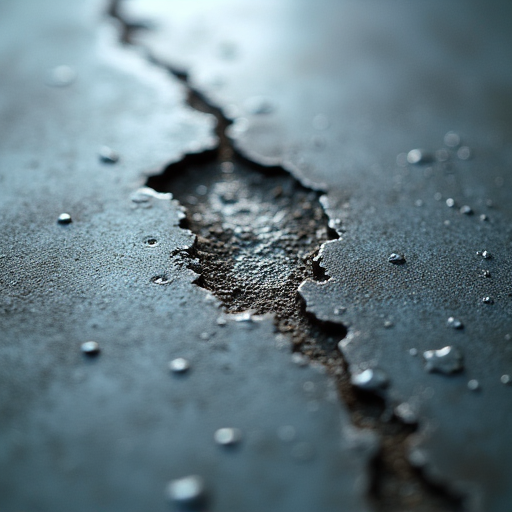}}
    \subcaptionbox{400}{\includegraphics[width=0.15\textwidth]{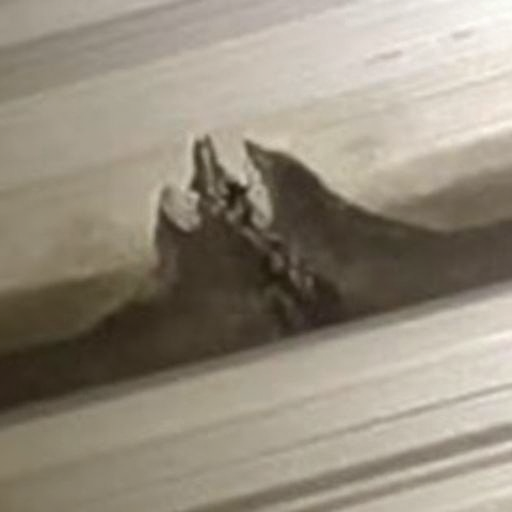}}
    \subcaptionbox{800}{\includegraphics[width=0.15\textwidth]{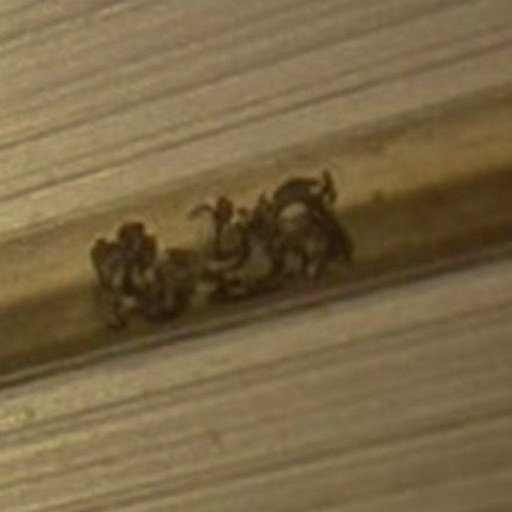}}
    \subcaptionbox{1{,}200}{\includegraphics[width=0.15\textwidth]{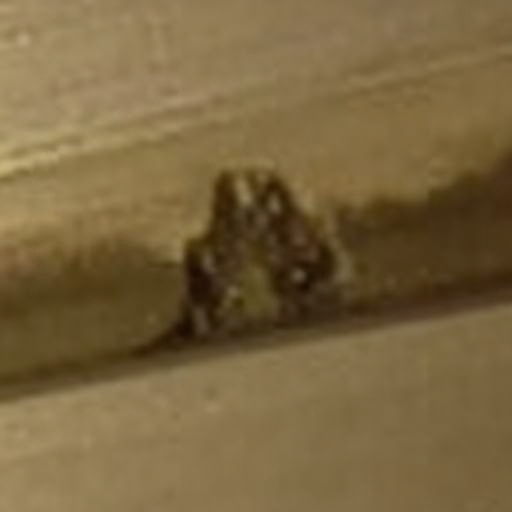}}
    \subcaptionbox{1{,}600}{\includegraphics[width=0.15\textwidth]{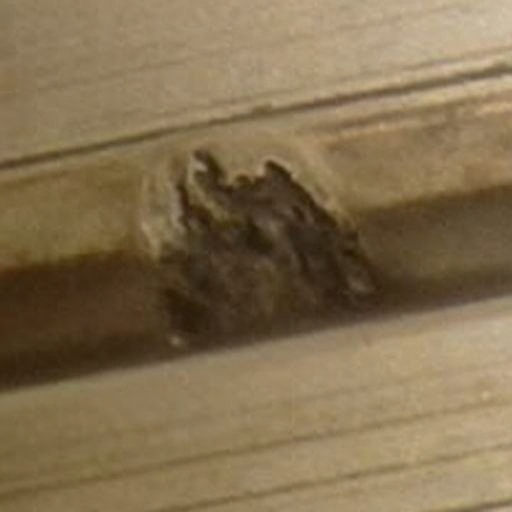}}
    \subcaptionbox{2{,}000}{\includegraphics[width=0.15\textwidth]{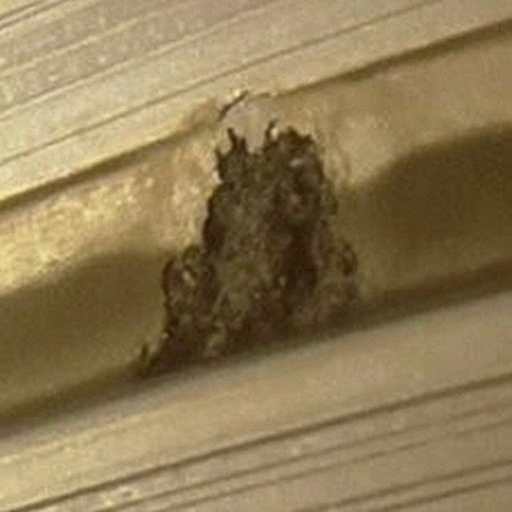}}
\caption{Prompt-fixed qualitative checkpoints for the random LoRA~\cite{hu2021loralowrankadaptationlarge} over $n$-training steps. Domain-consistent defect appearance emerges early \textbf{(b-e)} and stabilizes by 2{,}000 steps \textbf{(f)}.}
 \label{fig:fluxex4}
\end{figure}

\noindent\textbf{LoRA convergence over training steps.}
Fig.~\ref{fig:fluxex4} illustrates the evolution of the
random LoRA~\cite{hu2021loralowrankadaptationlarge} during training.
We generate a sample from a fixed prompt every 400 steps.
After the first checkpoint, the model already begins to reproduce
the target defect appearance more faithfully. By 2{,}000 steps,
the outputs exhibit defect morphology and background texture
visually consistent with the training data, indicating that
LoRA adaptation converges quickly in this domain. We observe a
similar trend for the scratch LoRA trained on
MSD~\cite{zhang2022fdsnet}.

\begin{figure}[!htbp]
    \centering
     \subcaptionbox{Simple}{\includegraphics[width=0.16\textwidth]{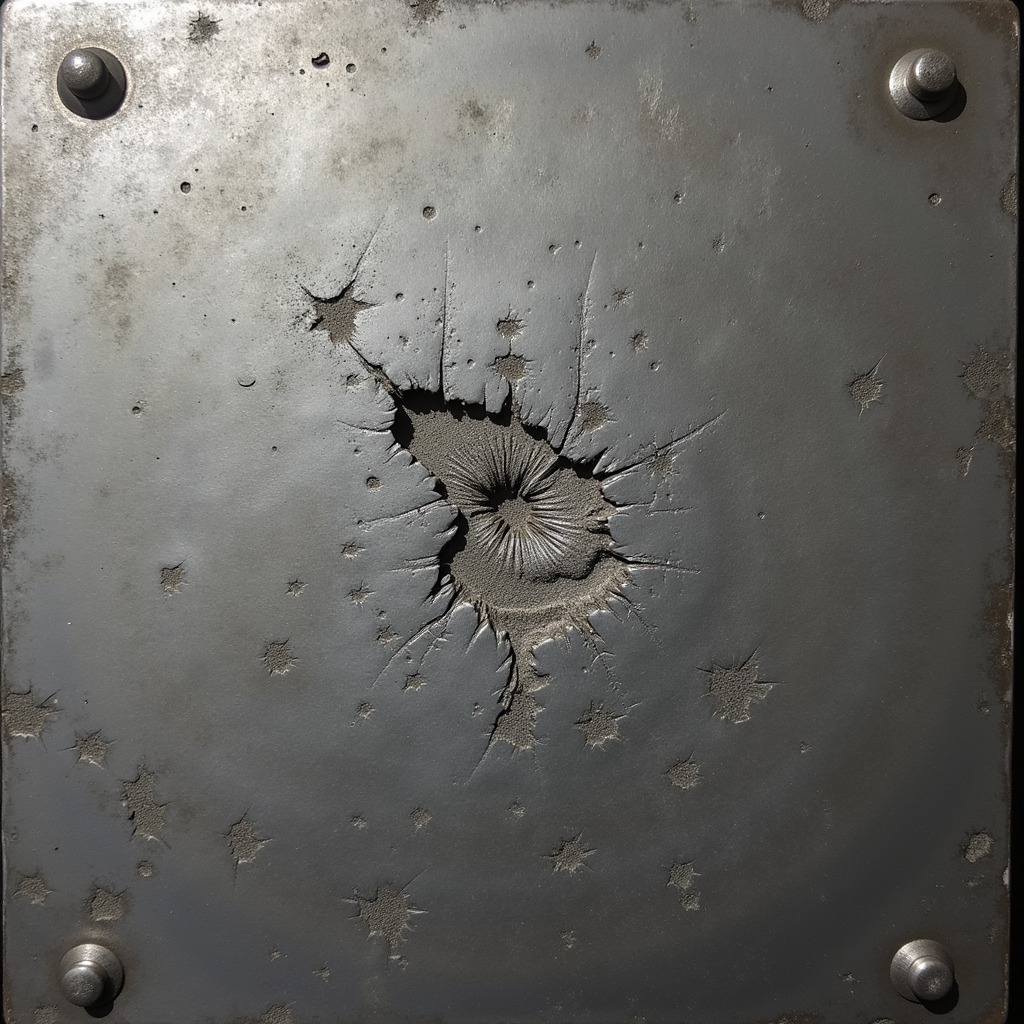}}
    \subcaptionbox{VLM}{\includegraphics[width=0.16\textwidth]{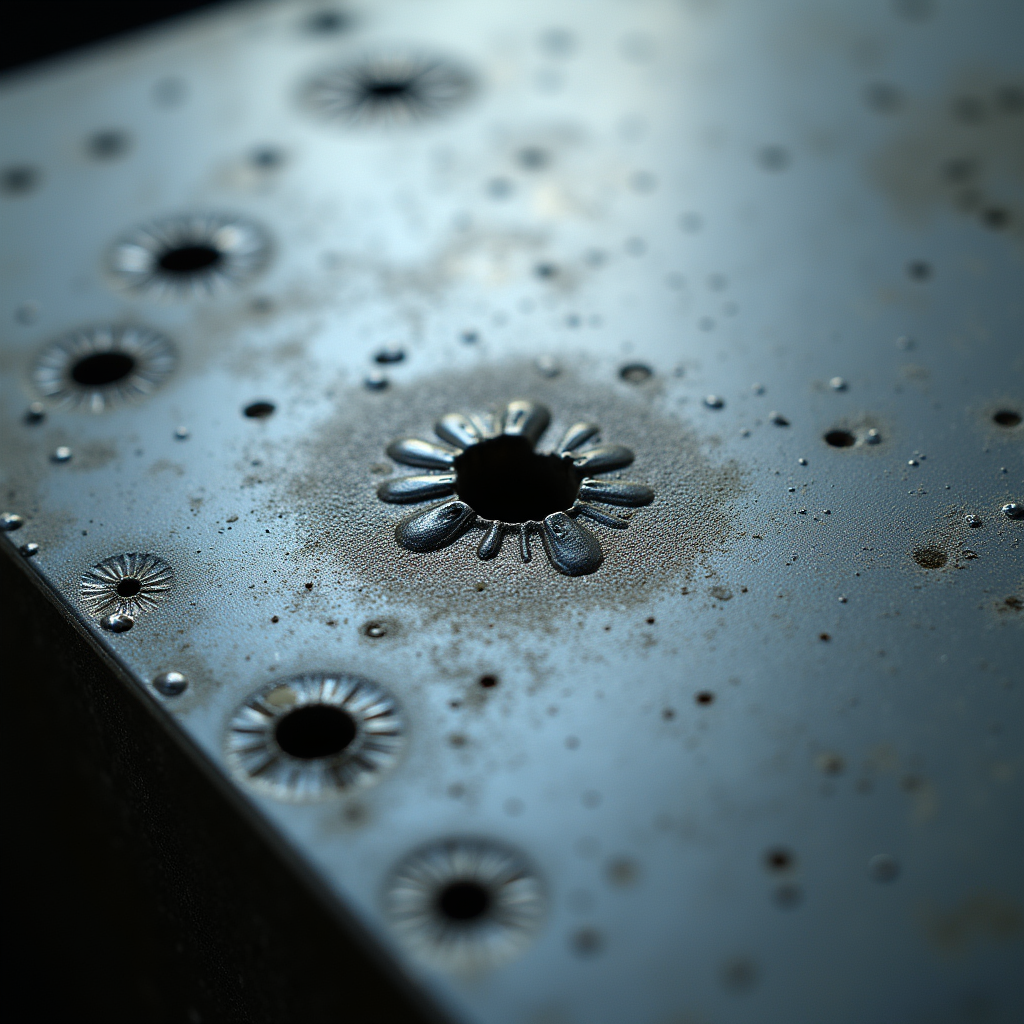}}
    \subcaptionbox{Image}{\includegraphics[width=0.16\textwidth]{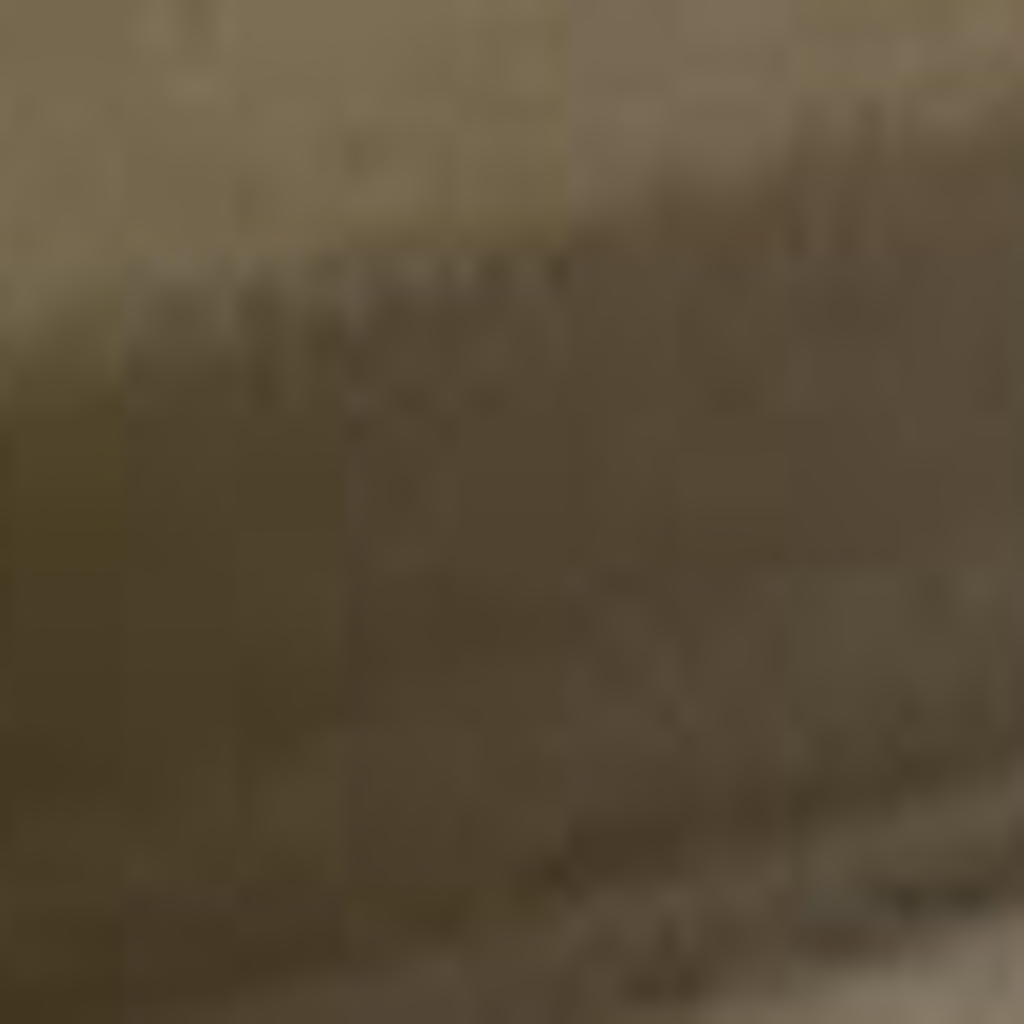}}
    \subcaptionbox{Mask}{\includegraphics[width=0.16\textwidth]{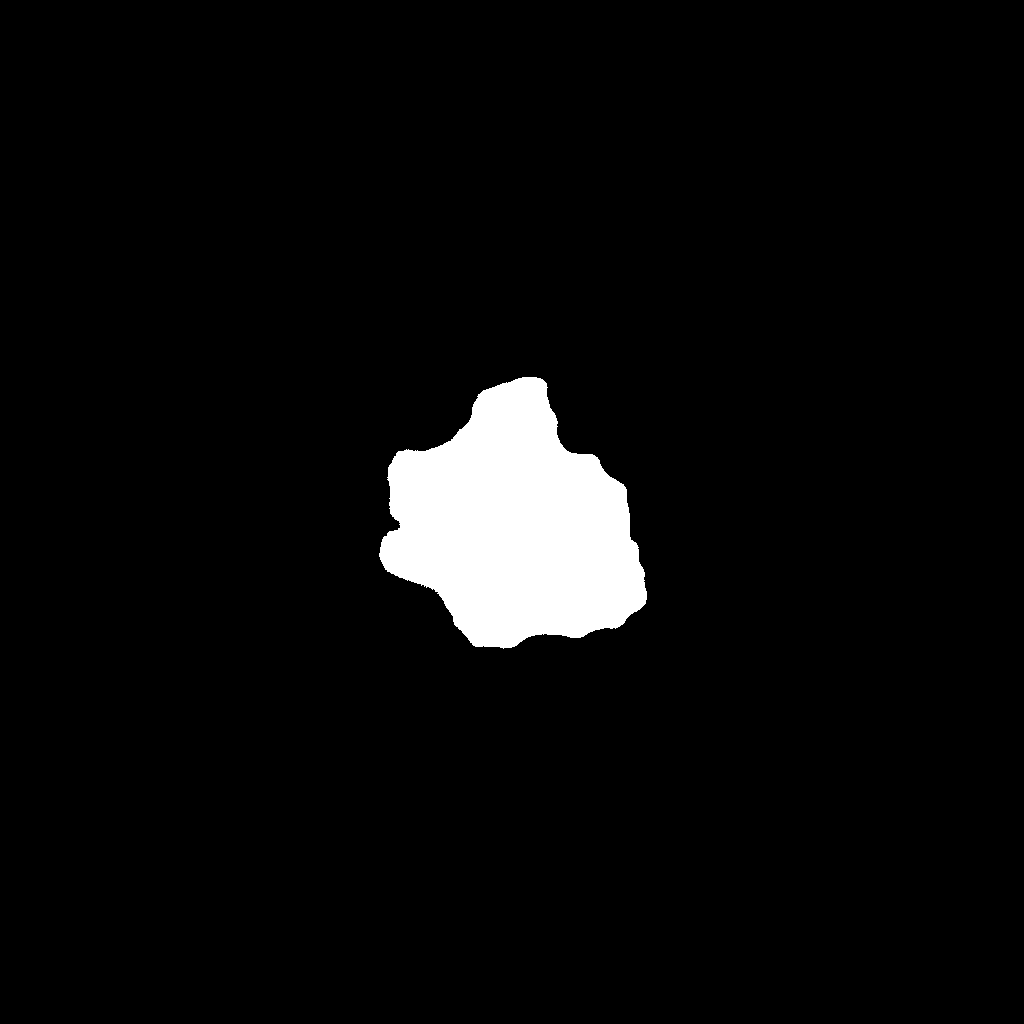}}
    \subcaptionbox{Simple}{\includegraphics[width=0.16\textwidth]{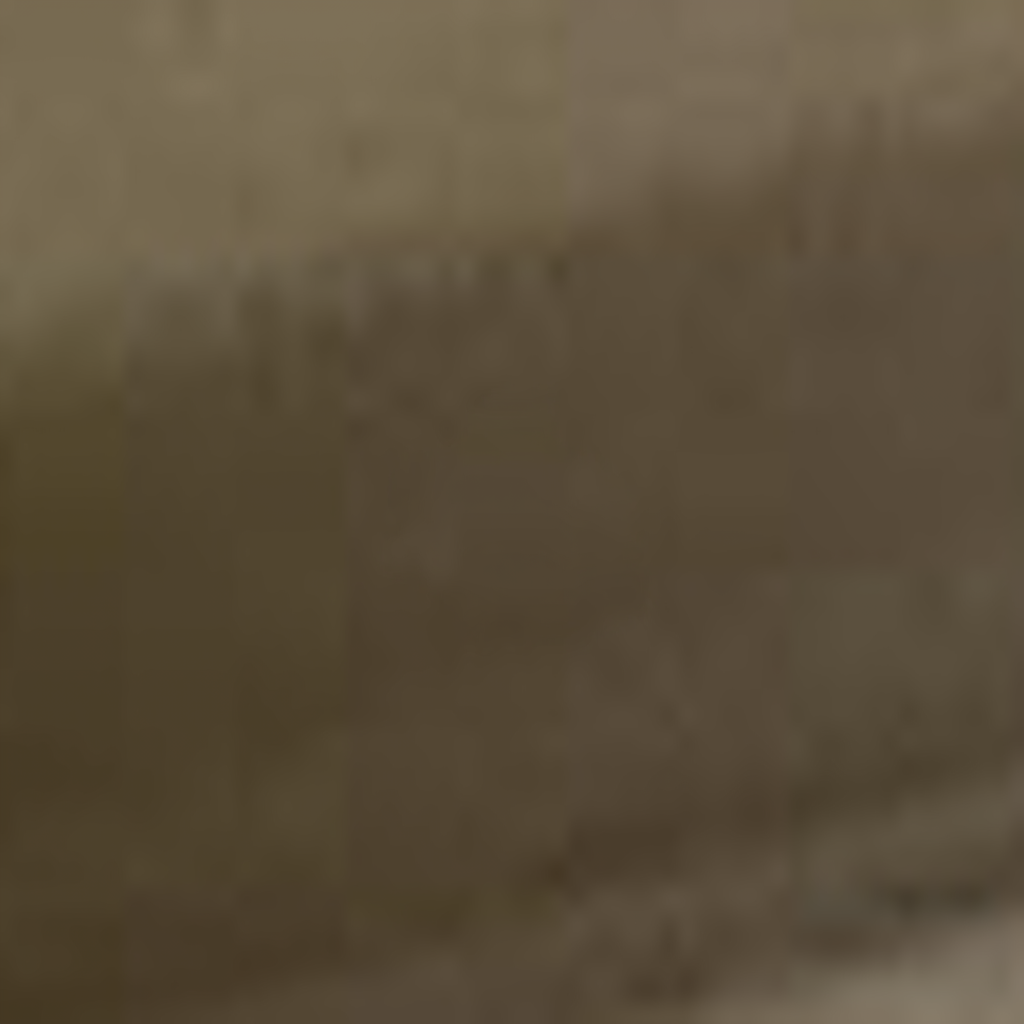}}
    \subcaptionbox{VLM}{\includegraphics[width=0.16\textwidth]{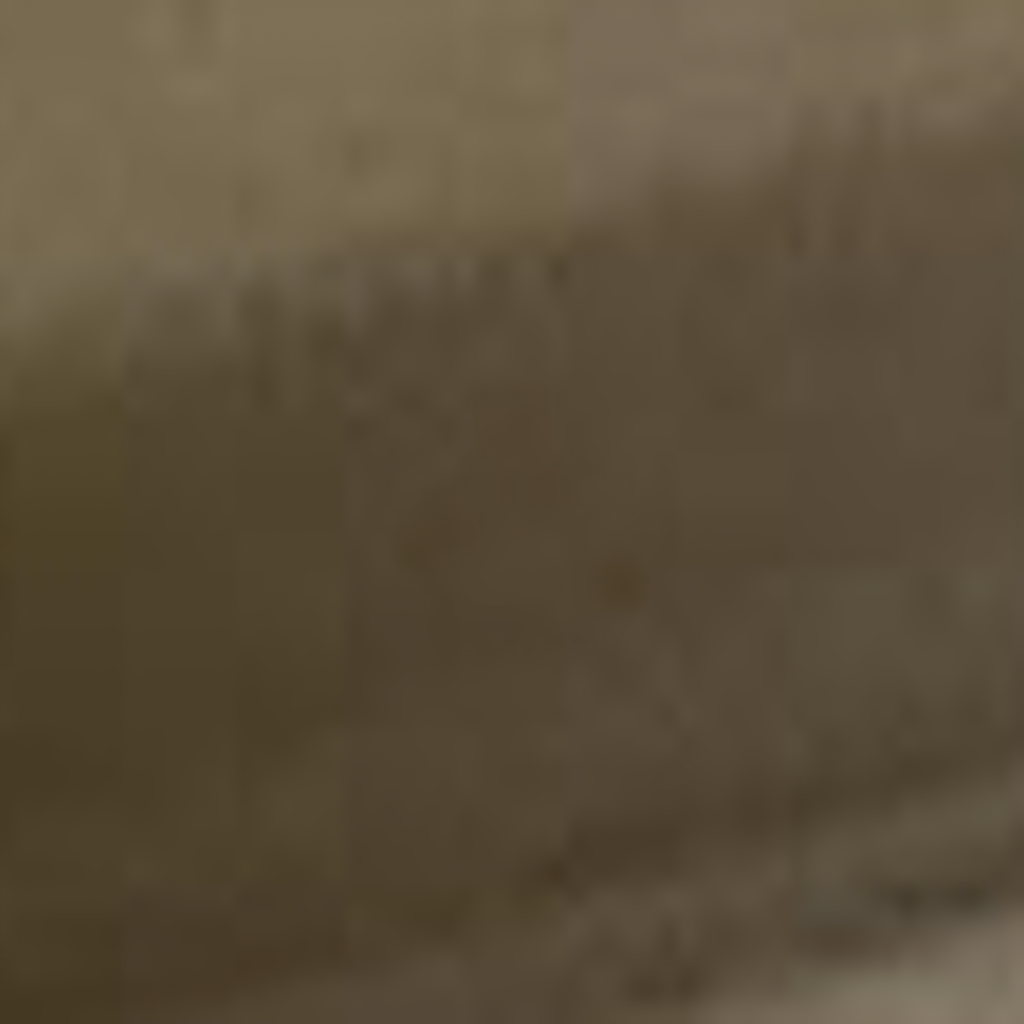}}
    \caption{Limitations of Flux.1-dev~\cite{flux2024,labs2025flux1kontextflowmatching} 
            without finetuning. 
            \textit{(a--b)}~Unconditional generations: prompt optimization alone fails to produce 
            domain-consistent defect appearances for BSData. 
            \textit{(c--f)}~Inpainting without LoRA~\cite{hu2021loralowrankadaptationlarge} adaptation: 
            given the same image patch and mask~\textit{(c--d)}, Flux.1-dev yields domain-inconsistent 
            defect structures regardless of the prompt used~\textit{(e--f)}. 
            ``VLM'' denotes a detailed prompt derived from a vision-language model, 
            whereas ``simple'' denotes a short, manually written sentence.}
 \label{fig:fluxex3}
\end{figure}

\noindent\textbf{Synthesis without LoRA}
Fig.~\ref{fig:fluxex3} \textbf{(a-b)} show unconditional generations from pretrained
Flux.1-dev~\cite{labs2025flux1kontextflowmatching,flux2024} without LoRA~\cite{hu2021loralowrankadaptationlarge}
finetuning. We compare a short simple prompt (\textit{``pitting defect on steel
surface''}) with the prompt derived with the VLM. Although both
outputs exhibit strong visual detail, neither resembles realistic industrial
surface defects. In particular, the longer prompt tends to produce more
imaginative and visually rich structures that do not match the texture and
defect morphology observed in BSData~\cite{schlagenhauf2021industrial}. This
suggests that prompt optimization alone is insufficient to align the model with
the target industrial domain.
Fig.~\ref{fig:fluxex3} \textit{(c)} shows a defect-free
image patch, the corresponding synthetic mask patch \textit{(d)}, and the inpainting outputs
for both prompts \textit{(e-f)}. In this setting, the generated region often lacks a plausible
defect structure or appears visually inconsistent with the surrounding surface.
These examples show that, without domain adaptation, the inpainting model does
not reproduce the appearance of the industrial dataset reliably, which
motivates LoRA~\cite{hu2021loralowrankadaptationlarge} fine-tuning as a necessary component
for defect synthesis.

\subsection{Detection performance YOLOX}\label{sec:appendix_yolox}
Tab.~\ref{tab:test_yolox} summarizes the
YOLOX-(S)~\cite{ge2021yolox} results on BSData.
The real-only baseline reaches $\text{AP}=0.595\pm0.020$.
Mixed regimes degrade monotonically with increasing synthetic
fraction, with the 75/25 split being the least harmful
($\text{AP}=0.583\pm0.005$), and synthetic-only training
collapsing to $\text{AP}=0.294\pm0.018$.
The union setting at 300 epochs (100\%\,R\,+\,100\%\,S) yields
the strongest overall result, surpassing the real-only baseline
across all metrics: $\text{AP}=0.606\pm0.013$,
$\text{AP50}=0.882\pm0.024$, and $\text{AR100}=0.643\pm0.012$.
Doubling the synthetic budget (100\%\,R\,+\,200\%\,S) does not
sustain these gains, with performance falling back toward the
baseline regardless of training duration. These results are
consistent with the YOLOv26 findings and further support the
conclusion that synthetic data is most effective as augmentation
when the full real set is retained, and that beyond a certain
synthetic volume diminishing returns set in.
\begin{table}[!htbp]
\scriptsize
\centering
\sisetup{separate-uncertainty=true}
\begin{tabularx}{\textwidth}{
    >{\raggedright\arraybackslash}X
    S[table-format=3.0, table-column-width=1.5cm]
    S[table-format=1.3(3), table-column-width=2.2cm]
    S[table-format=1.3(3), table-column-width=2.2cm]
    S[table-format=1.3(3), table-column-width=2.2cm]
    S[table-format=1.3(3), table-column-width=2.2cm]
}
\toprule
\textbf{Exp.(R/S)}
& \multicolumn{1}{c}{\textbf{Epochs}}
& \multicolumn{1}{c}{\textbf{AP~\up}}
& \multicolumn{1}{c}{\textbf{AP50~\up}}
& \multicolumn{1}{c}{\textbf{AP75~\up}}
& \multicolumn{1}{c}{\textbf{AR100~\up}} \\
\midrule
100/0
& 300 & \cellcolor{blue!15}\num{0.595 \pm 0.020} & \cellcolor{blue!15} \num{0.865 \pm 0.015} &  \cellcolor{blue!15}{ 0.685 $\pm$ 0.024} & \cellcolor{blue!15} \num{0.631 \pm 0.018} \\
75/25
& 300 & \num{0.583 \pm 0.005} & \num{0.855 \pm 0.022} & \cellcolor{yellow!50} {\bfseries 0.684 $\pm$ 0.043} & \num{0.620 \pm 0.004} \\
50/50
& 300 & \num{0.568 \pm 0.015} & \num{0.841 \pm 0.018} & \num{0.620 \pm 0.026} & \num{0.614 \pm 0.011} \\
25/75
& 300 & \num{0.536 \pm 0.013} & \num{0.825 \pm 0.007} & \num{0.630 \pm 0.069} & \num{0.590 \pm 0.022} \\
0/100
& 300 & \num{0.294 \pm 0.018} & \num{0.730 \pm 0.078} & \num{0.148 \pm 0.065} & \num{0.407 \pm 0.016} \\
100/100
& 150 & \cellcolor{gray!40}\num{0.589 \pm 0.003} & \num{0.859 \pm 0.022} & \num{0.637 \pm 0.020} & \cellcolor{gray!40}\num{0.634 \pm 0.008} \\
100/100
& 300 & \cellcolor{yellow!50} {\bfseries 0.606 $\pm$ 0.013} &  \cellcolor{yellow!50}{\bfseries 0.882 $\pm$ 0.024} & \num{0.646 \pm 0.044} &  \cellcolor{yellow!50}{\bfseries 0.643 $\pm$ 0.012} \\
100/200
& 100 & \num{0.578 \pm 0.002} & \cellcolor{gray!40}\num{0.874 \pm 0.007} & \cellcolor{gray!40}\num{0.664 \pm 0.017} & \num{0.628 \pm 0.002} \\
100/200
& 300 & \num{0.588 \pm 0.004} & \num{0.867 \pm 0.028} & \num{0.626 \pm 0.032} & \num{0.628 \pm 0.004} \\
\bottomrule
\end{tabularx}
\caption{YOLOX~\cite{ge2021yolox} results on the 
BSData~\cite{schlagenhauf2021industrial} test set (mean $\pm$ std over 3 runs). 
\emph{R} and \emph{S} denote the amount of real and synthetic training data relative to the full BSData training split, in \%. }
\label{tab:test_yolox}
\end{table}


 
\section{Cross-Dataset Transfer and Qualitative Ablations}
\begin{figure}[ht]
    \centering
    \includegraphics[width=1.\textwidth]{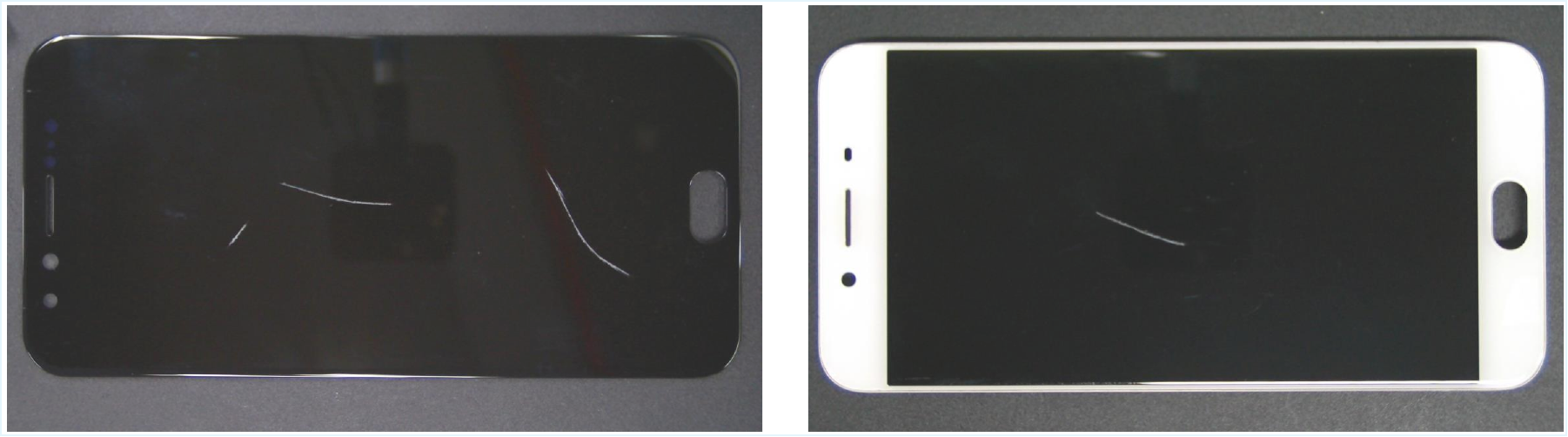}
   \caption{Data samples with multiple scratches \textit{(left)}
  and a single scratch \textit{(right)}.}
  \label{fig:scratches_msd_aax}
\end{figure}

\noindent\textbf{Data samples MSD.}
Fig.~\ref{fig:scratches_msd_aax} shows representative MSD samples illustrating the variability in scratch count and orientation across the dataset.

\begin{figure}[!htbp]
    \centering
    \includegraphics[width=0.49\textwidth]{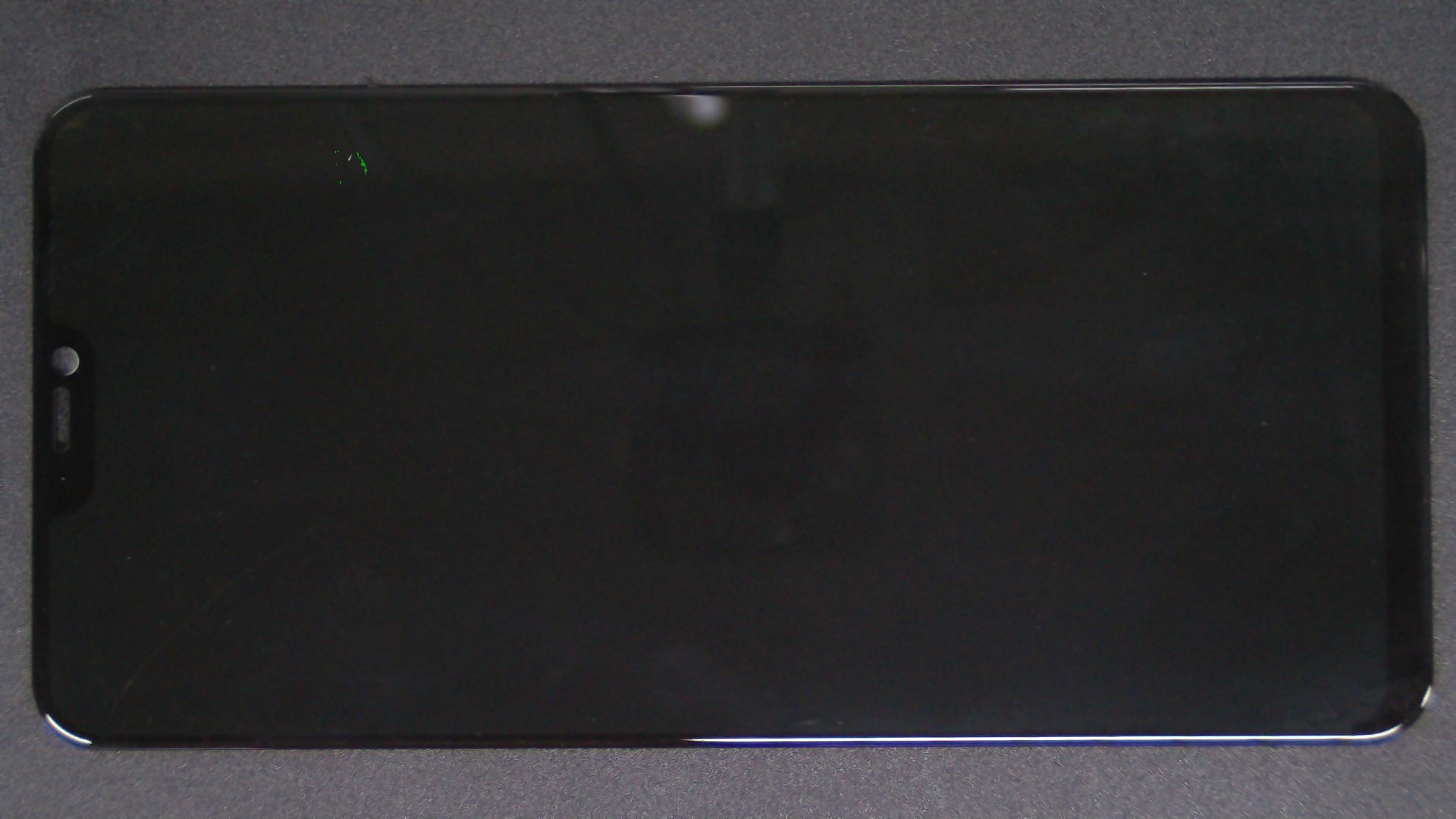}
    \hfill
    \includegraphics[width=0.49\textwidth]{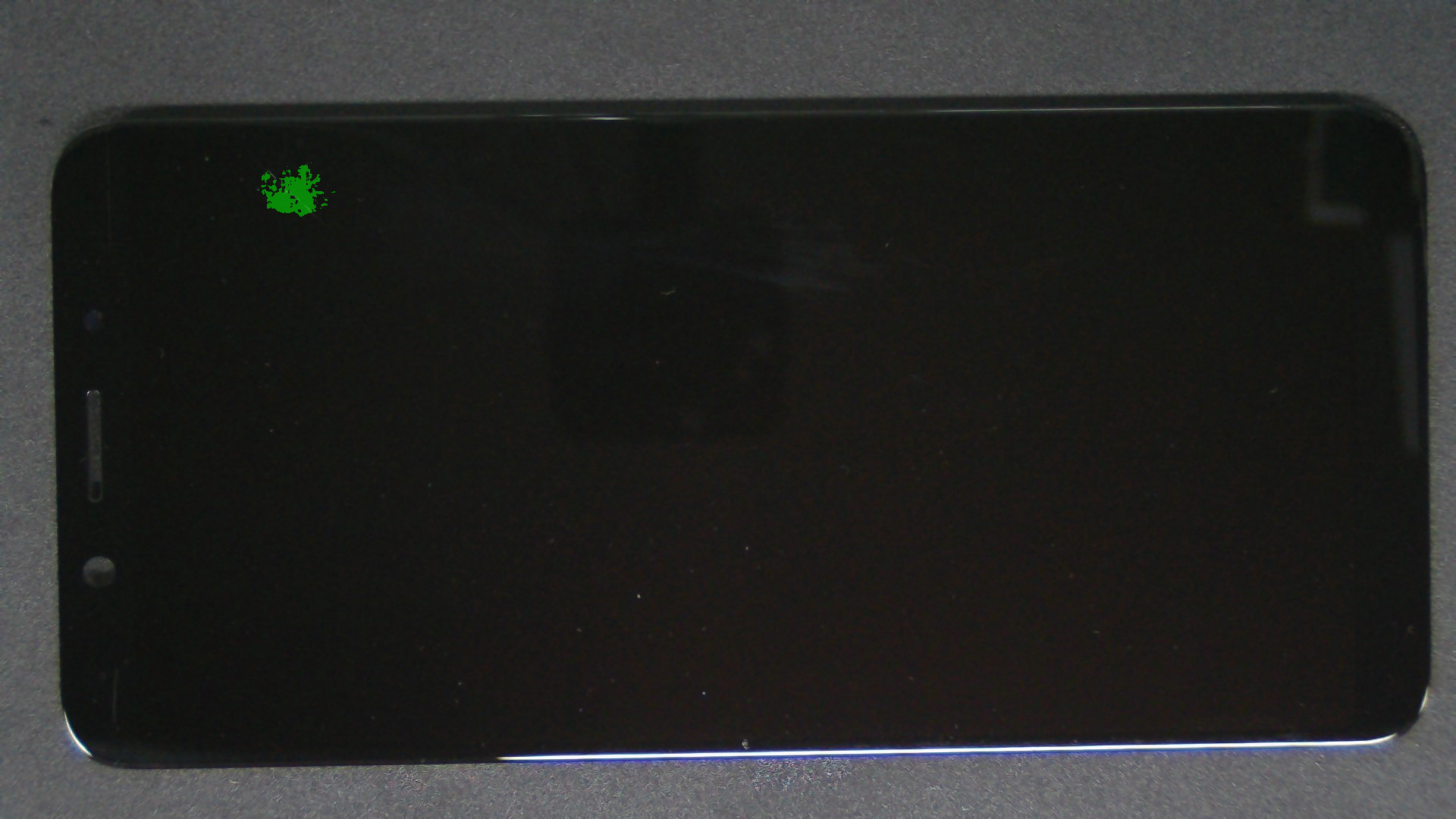}
    \medskip
    \includegraphics[width=0.49\textwidth]{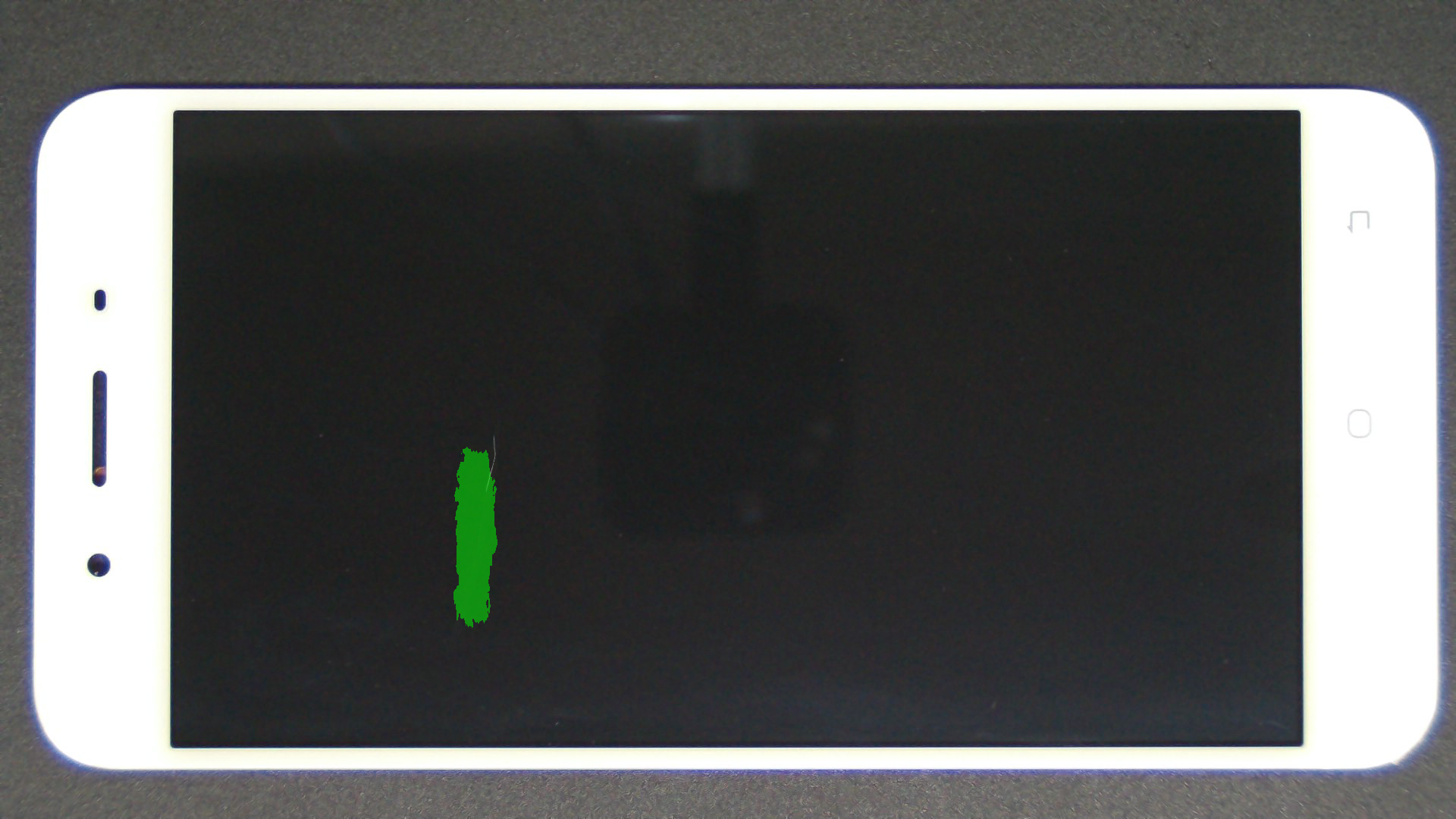}
    \hfill
    \includegraphics[width=0.49\textwidth]{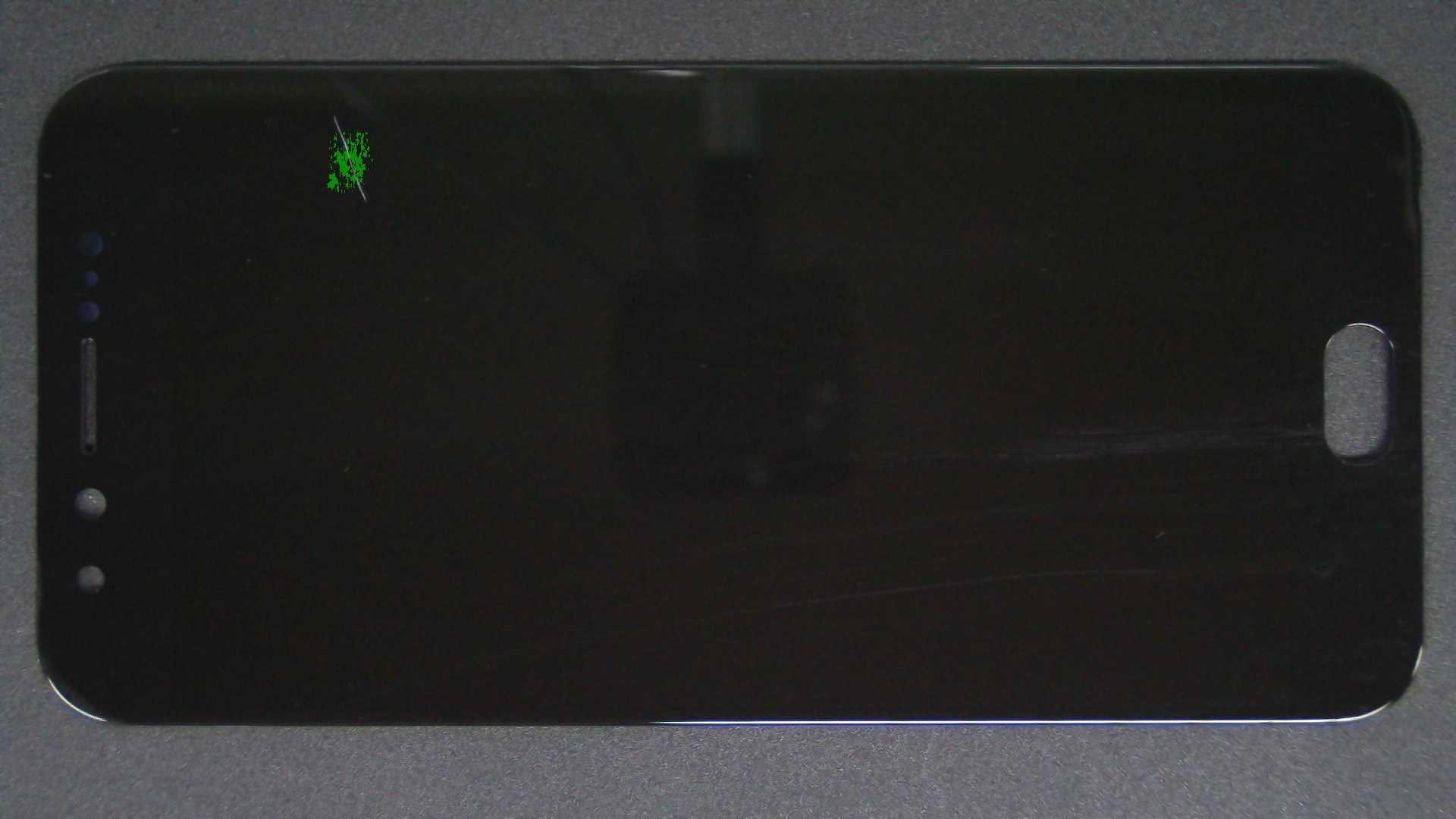}
\caption{Examples of failed SAM3~\cite{carion2025sam3segmentconcepts} segmentation
on synthetic MSD~\cite{zhang2022fdsnet} scratches. Predicted masks are
frequently displaced, diffuse, or substantially larger than the actual defect.}
 \label{fig:samfail}
\end{figure}

\noindent\textbf{Failure of SAM3 segmentation (MSD).}
During cross-dataset experiments on MSD~\cite{zhang2022fdsnet}, we
evaluated SAM~3~\cite{carion2025sam3segmentconcepts} as an automatic
segmentation component for synthetic scratch defects. Across a range
of reasonable settings, however, it failed to produce sufficiently
accurate masks. This is consistent with the visual properties of MSD
scratches, which are typically thin, low-contrast, and only weakly
textured. Consequently, predicted regions were often spatially
misaligned, overly diffuse, or substantially larger than the true
defect extent. Representative failure cases are shown in
Fig.~\ref{fig:samfail}.

\FloatBarrier

These observations suggest that robust scratch segmentation in this
setting requires careful calibration of thresholds and prompts. While
such tuning may be feasible for individual cases, it does not scale
to automatic large-scale dataset generation. We therefore resort to
synthetic inpainting masks as surrogate annotations for
MSD~\cite{zhang2022fdsnet}. Although these masks tend to overestimate
the defect extent, they provide a sufficient basis for deriving
bounding boxes and maintaining consistent training labels.

\begin{figure}[!htbp]
\centering
\fbox{%
  \begin{minipage}{0.96\textwidth}
  \small
  \textbf{Resulting prompt:}\\[2pt]
  \textit{high contrast scratch defect on dark glass display, thin linear scratch, occasional diagonal orientation, sharp edges, isolated single defect, reflective glossy surface with subtle metallic sheen, fine texture on smooth surface, close-up industrial inspection photo, uniform lighting with faint glow, minimal dark background, minimal noise shallow depth of field}
  \end{minipage}%
}
\caption{Final MSD~\cite{zhang2022fdsnet} prompt derived from frequent Qwen2-VL~\cite{wang2024qwen2vl} 
tags and light manual pruning. 
The prompt emphasizes scratch geometry, reflective display appearance, and controlled acquisition conditions.}
\label{fig:prompt_qwen_summary_msd}
\end{figure}

\bibliographystyle{plain}
\bibliography{main}

@String(CVPR  = {IEEE Conf. Comput. Vis. Pattern Recog.})

@String(ICCV  = {Int. Conf. Comput. Vis.})

@String(ICML  = {Int. Conf. Mach. Learn.})

@String(CVPRW = {IEEE Conf. Comput. Vis. Pattern Recog. Worksh.})

@String(ICASSP=	{Int. Conf. on Acoustics, Speech and Sig. Proc.})

@String(CVPR  = {CVPR})

@String(ICCV  = {ICCV})

@String(ICML  = {ICML})

@String(CVPRW = {CVPRW})

@String(ICASSP=	{ICASSP})

@online{azizi2023synthetic,
  title       = {Synthetic Data from Diffusion Models Improves ImageNet Classification},
  author      = {Azizi, Shekoofeh and Kornblith, Simon and Saharia, Chitwan and Norouzi, Mohammad and Fleet, David J.},
  date        = {2023},
  eprint      = {2304.08466},
  eprinttype  = {arxiv},
  eprintclass = {cs.CV},
  url         = {https://arxiv.org/abs/2304.08466},
  urldate     = {2026-01-26},
}

@online{nguyen2025we,
  title       = {Do We Need All the Synthetic Data? Targeted Synthetic Image Augmentation via Diffusion Models},
  author      = {Nguyen, Dang and Li, Jiping and Zheng, Jinghao and Mirzasoleiman, Baharan},
  date        = {2025},
  eprint      = {2505.21574},
  eprinttype  = {arxiv},
  eprintclass = {cs.CV},
  url         = {https://arxiv.org/abs/2505.21574},
  urldate     = {2026-01-26},
}

@inproceedings{tayeb2025defectdiffusion,
  title     = {DefectDiffusion: A Generative Diffusion Model for Robust Data Augmentation in Industrial Defect Detection},
  author    = {Tayeb, Adnan Md and Nakayiza, Hope Leticia and Shin, Heejae and Lee, Seungmin and Lee, Jae-Min and Kim, Dong-Seong},
  booktitle = {2025 International Conference on Artificial Intelligence in Information and Communication (ICAIIC)},
  pages     = {66--71},
  date      = {2025},
  doi       = {10.1109/ICAIIC64266.2025.10920732},
  url       = {https://doi.org/10.1109/ICAIIC64266.2025.10920732},
}

@article{xu2024systematic,
  title        = {A systematic review and evaluation of synthetic simulated data generation strategies for deep learning applications in construction},
  author       = {Xu, Liqun and Liu, Hexu and Xiao, Bo and Luo, Xiaowei and Veeramani, Dharmaraj and Zhu, Zhenhua},
  journaltitle = {Advanced Engineering Informatics},
  volume       = {62},
  eid          = {102699},
  date         = {2024-10},
  doi          = {10.1016/j.aei.2024.102699},
  url          = {https://doi.org/10.1016/j.aei.2024.102699},
}

@online{ge2021yolox,
  title       = {{YOLOX}: Exceeding {YOLO} Series in 2021},
  author      = {Ge, Zheng and Liu, Songtao and Wang, Feng and Li, Zeming and Sun, Jian},
  date        = {2021},
  eprint      = {2107.08430},
  eprinttype  = {arxiv},
  eprintclass = {cs.CV},
  url         = {https://arxiv.org/abs/2107.08430},
  urldate     = {2026-01-26},
}

@online{hessel2021clipscore,
  title       = {{CLIPScore}: A Reference-free Evaluation Metric for Image Captioning},
  author      = {Hessel, Jack and Holtzman, Ari and Forbes, Maxwell and Le Bras, Ronan and Choi, Yejin},
  date        = {2021},
  eprint      = {2104.08718},
  eprinttype  = {arxiv},
  eprintclass = {cs.CV},
  url         = {https://arxiv.org/abs/2104.08718},
  urldate     = {2026-01-26},
}

@software{flux2024,
  author       = {{Black Forest Labs}},
  title        = {{FLUX}},
  date         = {2024},
  url          = {https://github.com/black-forest-labs/flux},
  urldate      = {2026-01-26},
}

@article{schlagenhauf2021industrial,
  title        = {Industrial Machine Tool Component Surface Defect Dataset},
  author       = {Schlagenhauf, Tobias and Landwehr, Magnus},
  journaltitle = {Data in Brief},
  volume       = {39},
  eid          = {107643},
  date         = {2021-12},
  doi          = {10.1016/j.dib.2021.107643},
  url          = {https://doi.org/10.1016/j.dib.2021.107643},
}

@article{jain2022synthetic,
  title        = {Synthetic data augmentation for surface defect detection and classification using deep learning},
  author       = {Jain, Saksham and Seth, Gautam and Paruthi, Arpit and Soni, Umang and Kumar, G.},
  journaltitle = {Journal of Intelligent Manufacturing},
  volume       = {33},
  date         = {2022-04},
  doi          = {10.1007/s10845-020-01710-x},
  url          = {https://doi.org/10.1007/s10845-020-01710-x},
}

@incollection{lebert2022synthetic,
  title     = {Synthetic Data Generation for Surface Defect Detection},
  author    = {Lebert, D{\'e}borah and Plouzeau, J{\'e}r{\'e}my and Farrugia, Jean-Philippe and Danglade, Florence and Merienne, Fr{\'e}d{\'e}ric},
  booktitle = {Extended Reality},
  publisher = {Springer Nature Switzerland},
  location  = {Cham},
  pages     = {198--208},
  date      = {2022},
  doi       = {10.1007/978-3-031-15553-6_15},
  url       = {https://doi.org/10.1007/978-3-031-15553-6_15},
  isbn      = {978-3-031-15553-6},
}

@article{liu2024defectgan,
  title        = {DefectGAN: Synthetic Data Generation for EMU Defects Detection with Limited Data},
  author       = {Liu, Scarlett and Ni, Hai and Li, Chao and Zou, Yukang and Luo, Yiping},
  journaltitle = {IEEE Sensors Journal},
  volume       = {24},
  number       = {11},
  pages        = {17638--17652},
  date         = {2024},
  doi          = {10.1109/JSEN.2024.3386711},
}

@online{jin2022survey,
  title       = {A Survey of Surface Defect Detection of Industrial Products Based on a Small Number of Labeled Data},
  author      = {Jin, Qifan and Chen, Li},
  date        = {2022},
  eprint      = {2203.05733},
  eprinttype  = {arxiv},
  eprintclass = {cs.CV},
  url         = {https://arxiv.org/abs/2203.05733},
  urldate     = {2026-01-26},
}

@inproceedings{tayeb2024defectgen,
  title     = {{DefectGen}: Few-Shot Defect Image Generation using Stable Diffusion for Steel Surface Analysis},
  author    = {Tayeb, Adnan Md and Nakayiza, Hope Leticia and Shin, Heejae and Lee, Seungmin and Lee, Chaesoo and Lee, YeongHun and Kim, Dong-Seong and Lee, Jae-Min},
  booktitle = {2024 15th International Conference on Information and Communication Technology Convergence (ICTC)},
  date      = {2024},
  pages     = {2087--2092},
  doi       = {10.1109/ICTC62082.2024.10827273},
  url       = {https://doi.org/10.1109/ICTC62082.2024.10827273},
}

@online{song2025defectfill,
  title       = {{DefectFill}: Realistic Defect Generation with Inpainting Diffusion Model for Visual Inspection},
  author      = {Song, Jaewoo and Park, Daemin and Baek, Kanghyun and Lee, Sangyub and Choi, Jooyoung and Kim, Eunji and Yoon, Sungroh},
  date        = {2025},
  eprint      = {2503.13985},
  eprinttype  = {arxiv},
  eprintclass = {cs.CV},
  url         = {https://arxiv.org/abs/2503.13985},
  urldate     = {2026-01-26},
}

@article{ali2024anomalycontrol,
  title        = {{AnomalyControl}: Few-Shot Anomaly Generation by {ControlNet} Inpainting},
  author       = {Ali, Musawar and Fioraio, Nicola and Salti, Samuele and Di Stefano, Luigi},
  journaltitle = {IEEE Access},
  volume       = {12},
  pages        = {192903--192914},
  date         = {2024},
  doi          = {10.1109/ACCESS.2024.3520002},
  url          = {https://doi.org/10.1109/ACCESS.2024.3520002},
}

@online{chen2024lwdetrtransformerreplacementyolo,
  title       = {{LW-DETR}: A Transformer Replacement to {YOLO} for Real-Time Detection},
  author      = {Chen, Qiang and Su, Xiangbo and Zhang, Xinyu and Wang, Jian and Chen, Jiahui and Shen, Yunpeng and Han, Chuchu and Chen, Ziliang and Xu, Weixiang and Li, Fanrong and Zhang, Shan and Yao, Kun and Ding, Errui and Zhang, Gang and Wang, Jingdong},
  date        = {2024},
  eprint      = {2406.03459},
  eprinttype  = {arxiv},
  eprintclass = {cs.CV},
  url         = {https://arxiv.org/abs/2406.03459},
  urldate     = {2026-01-26},
}

@online{hu2021loralowrankadaptationlarge,
  title       = {{LoRA}: Low-Rank Adaptation of Large Language Models},
  author      = {Hu, Edward J. and Shen, Yelong and Wallis, Phillip and Allen-Zhu, Zeyuan and Li, Yuanzhi and Wang, Shean and Lu, Wang and Chen, Weizhu},
  date        = {2021},
  eprint      = {2106.09685},
  eprinttype  = {arxiv},
  eprintclass = {cs.CL},
  url         = {https://arxiv.org/abs/2106.09685},
  urldate     = {2026-01-26},
}

@online{yang2025qwen3technicalreport,
  title       = {{Qwen3} Technical Report},
  author      = {Yang, An and Li, Anfeng and Yang, Baosong and Zhang, Beichen and Hui, Binyuan and Zheng, Bo and Yu, Bowen and Gao, Chang and Huang, Chengen and Lv, Chenxu and Zheng, Chujie and Liu, Dayiheng and Zhou, Fan and Huang, Fei and Hu, Feng and Ge, Hao and Wei, Haoran and Lin, Huan and Tang, Jialong and Yang, Jian and Tu, Jianhong and Zhang, Jianwei and Yang, Jianxin and Yang, Jiaxi and Zhou, Jing and Zhou, Jingren and Lin, Junyang and Dang, Kai and Bao, Keqin and Yang, Kexin and Yu, Le and Deng, Lianghao and Li, Mei and Xue, Mingfeng and Li, Mingze and Zhang, Pei and Wang, Peng and Zhu, Qin and Men, Rui and Gao, Ruize and Liu, Shixuan and Luo, Shuang and Li, Tianhao and Tang, Tianyi and Yin, Wenbiao and Ren, Xingzhang and Wang, Xinyu and Zhang, Xinyu and Ren, Xuancheng and Fan, Yang and Su, Yang and Zhang, Yichang and Zhang, Yinger and Wan, Yu and Liu, Yuqiong and Wang, Zekun and Cui, Zeyu and Zhang, Zhenru and Zhou, Zhipeng and Qiu, Zihan},
  date        = {2025},
  eprint      = {2505.09388},
  eprinttype  = {arxiv},
  eprintclass = {cs.CL},
  url         = {https://arxiv.org/abs/2505.09388},
  urldate     = {2026-01-26},
}

@online{sapkota2026yolo26keyarchitecturalenhancements,
  title       = {{YOLO26}: Key Architectural Enhancements and Performance Benchmarking for Real-Time Object Detection},
  author      = {Sapkota, Ranjan and Cheppally, Rahul Harsha and Sharda, Ajay and Karkee, Manoj},
  date        = {2025},
  eprint      = {2509.25164},
  eprinttype  = {arxiv},
  eprintclass = {cs.CV},
  url         = {https://arxiv.org/abs/2509.25164},
  urldate     = {2026-01-26},
}

@inproceedings{zhang2022fdsnet,
  title     = {{FDSNeT}: An Accurate Real-Time Surface Defect Segmentation Network},
  author    = {Zhang, Jian and Ding, Runwei and Ban, Miaoju and Guo, Tianyu},
  booktitle = {ICASSP 2022 -- 2022 IEEE International Conference on Acoustics, Speech and Signal Processing (ICASSP)},
  date      = {2022},
  pages     = {3803--3807},
  doi       = {10.1109/ICASSP43922.2022.9747311},
  url       = {https://doi.org/10.1109/ICASSP43922.2022.9747311},
}

@misc{lv2023detrs,
      title={DETRs Beat YOLOs on Real-time Object Detection},
      author={Yian Zhao and Wenyu Lv and Shangliang Xu and Jinman Wei and Guanzhong Wang and Qingqing Dang and Yi Liu and Jie Chen},
      year={2023},
      eprint={2304.08069},
      archivePrefix={arXiv},
      primaryClass={cs.CV}
}

@misc{lv2024rtdetrv2improvedbaselinebagoffreebies,
      title={RT-DETRv2: Improved Baseline with Bag-of-Freebies for Real-Time Detection Transformer}, 
      author={Wenyu Lv and Yian Zhao and Qinyao Chang and Kui Huang and Guanzhong Wang and Yi Liu},
      year={2024},
      eprint={2407.17140},
      archivePrefix={arXiv},
      primaryClass={cs.CV},
      url={https://arxiv.org/abs/2407.17140}, 
}

@misc{rf-detr,
    title={RF-DETR: Neural Architecture Search for Real-Time Detection Transformers},
    author={Isaac Robinson and Peter Robicheaux and Matvei Popov and Deva Ramanan and Neehar Peri},
    year={2025},
    eprint={2511.09554},
    archivePrefix={arXiv},
    primaryClass={cs.CV},
    url={https://arxiv.org/abs/2511.09554},
}

@inproceedings{redmon2016yolo,
  author={Redmon, Joseph and Divvala, Santosh and Girshick, Ross and Farhadi, Ali},
  booktitle={2016 IEEE Conference on Computer Vision and Pattern Recognition (CVPR)}, 
  title={You Only Look Once: Unified, Real-Time Object Detection}, 
  year={2016},
  volume={},
  number={},
  pages={779-788},
  doi={10.1109/CVPR.2016.91}}

@misc{redmon2018yolov3,
      title={YOLOv3: An Incremental Improvement}, 
      author={Joseph Redmon and Ali Farhadi},
      year={2018},
      eprint={1804.02767},
      archivePrefix={arXiv},
      primaryClass={cs.CV},
      url={https://arxiv.org/abs/1804.02767}, 
}

@article{bochkovskiy2020yolov4,
  title={YOLOv4: Optimal Speed and Accuracy of Object Detection},
  author={Alexey Bochkovskiy and Chien-Yao Wang and Hong-Yuan Mark Liao},
  journal={ArXiv},
  year={2020},
  volume={abs/2004.10934},
  url={https://api.semanticscholar.org/CorpusID:216080778}
}

@article{carion2020detr,
  title={End-to-End Object Detection with Transformers},
  author={Nicolas Carion and Francisco Massa and Gabriel Synnaeve and Nicolas Usunier and Alexander Kirillov and Sergey Zagoruyko},
  journal={ArXiv},
  year={2020},
  volume={abs/2005.12872},
  url={https://api.semanticscholar.org/CorpusID:218889832}
}

@inproceedings{denninger2020blenderproc,
          author = {Denninger, Maximilian and Sundermeyer, Martin and Winkelbauer, Dominik and Olefir, Dmitry and Hodan, Tomas and Zidan, Youssef and Elbadrawy, Mohamad and Knauer, Markus and Katam, Harinandan and Lodhi, Ahsan},
           month = {Juli},
           title = {BlenderProc: Reducing the Reality Gap with Photorealistic Rendering},
            note = {Video presentation: https://www.youtube.com/watch?v=tQ59iGVnJWM},
       booktitle = {16th Robotics: Science and Systems, RSS 2020, Workshops},
            year = {2020},
             url = {https://elib.dlr.de/139317/},
            isbn = {978-0-9923747-6-1}
}

@inproceedings{tremblay2018training,
  author={Tremblay, Jonathan and Prakash, Aayush and Acuna, David and Brophy, Mark and Jampani, Varun and Anil, Cem and To, Thang and Cameracci, Eric and Boochoon, Shaad and Birchfield, Stan},
  booktitle={2018 IEEE/CVF Conference on Computer Vision and Pattern Recognition Workshops (CVPRW)}, 
  title={Training Deep Networks with Synthetic Data: Bridging the Reality Gap by Domain Randomization}, 
  year={2018},
  volume={},
  number={},
  pages={1082-10828},
  doi={10.1109/CVPRW.2018.00143}}

@misc{tremblay2018falling,
      title={Falling Things: A Synthetic Dataset for 3D Object Detection and Pose Estimation}, 
      author={Jonathan Tremblay and Thang To and Stan Birchfield},
      year={2018},
      eprint={1804.06534},
      archivePrefix={arXiv},
      primaryClass={cs.CV},
      url={https://arxiv.org/abs/1804.06534}, 
}

@misc{li2021cutpaste,
      title={CutPaste: Self-Supervised Learning for Anomaly Detection and Localization}, 
      author={Chun-Liang Li and Kihyuk Sohn and Jinsung Yoon and Tomas Pfister},
      year={2021},
      eprint={2104.04015},
      archivePrefix={arXiv},
      primaryClass={cs.CV},
      url={https://arxiv.org/abs/2104.04015}, 
}

@inproceedings{zavrtanik2021draem,
  author={Zavrtanik, Vitjan and Kristan, Matej and Skočaj, Danijel},
  booktitle={2021 IEEE/CVF International Conference on Computer Vision (ICCV)}, 
  title={DRÆM – A discriminatively trained reconstruction embedding for surface anomaly detection}, 
  year={2021},
  volume={},
  number={},
  pages={8310-8319},
  doi={10.1109/ICCV48922.2021.00822}}

@inproceedings{isola2017pix2pix,
 author={Tahmid, Marjana and Alam, Md. Samiul and Rao, Namratha and Ashrafi, Kazi Muhammad Asif},
  booktitle={2023 IEEE 9th International Women in Engineering (WIE) Conference on Electrical and Computer Engineering (WIECON-ECE)}, 
  title={Image-to-Image Translation with Conditional Adversarial Networks}, 
  year={2023},
  volume={},
  number={},
  pages={1-5},
  doi={10.1109/WIECON-ECE60392.2023.10456447}}

@inproceedings{zhu2017cyclegan,
  author={Zhu, Jun-Yan and Park, Taesung and Isola, Phillip and Efros, Alexei A.},
  booktitle={2017 IEEE International Conference on Computer Vision (ICCV)}, 
  title={Unpaired Image-to-Image Translation Using Cycle-Consistent Adversarial Networks}, 
  year={2017},
  volume={},
  number={},
  pages={2242-2251},
  doi={10.1109/ICCV.2017.244}}

@inproceedings{ho2020ddpm,
author = {Ho, Jonathan and Jain, Ajay and Abbeel, Pieter},
title = {Denoising diffusion probabilistic models},
year = {2020},
isbn = {9781713829546},
publisher = {Curran Associates Inc.},
address = {Red Hook, NY, USA},
booktitle = {Proceedings of the 34th International Conference on Neural Information Processing Systems},
articleno = {574},
numpages = {12},
location = {Vancouver, BC, Canada},
series = {NIPS '20}
}

@misc{rombach2022stablediffusion,
      title={High-Resolution Image Synthesis with Latent Diffusion Models}, 
      author={Robin Rombach and Andreas Blattmann and Dominik Lorenz and Patrick Esser and Björn Ommer},
      year={2022},
      eprint={2112.10752},
      archivePrefix={arXiv},
      primaryClass={cs.CV},
      url={https://arxiv.org/abs/2112.10752}, 
}

@inproceedings{dhariwal2021diffusionbeatsgans,
author = {Dhariwal, Prafulla and Nichol, Alex},
title = {Diffusion models beat GANs on image synthesis},
year = {2021},
isbn = {9781713845393},
publisher = {Curran Associates Inc.},
address = {Red Hook, NY, USA},
booktitle = {Proceedings of the 35th International Conference on Neural Information Processing Systems},
articleno = {672},
numpages = {15},
series = {NIPS '21}
}

@misc{zhang2023controlnet,
      title={Adding Conditional Control to Text-to-Image Diffusion Models}, 
      author={Lvmin Zhang and Anyi Rao and Maneesh Agrawala},
      year={2023},
      eprint={2302.05543},
      archivePrefix={arXiv},
      primaryClass={cs.CV},
      url={https://arxiv.org/abs/2302.05543}, 
}

@inproceedings{ruiz2023dreambooth,
  author={Ruiz, Nataniel and Li, Yuanzhen and Jampani, Varun and Pritch, Yael and Rubinstein, Michael and Aberman, Kfir},
  booktitle={2023 IEEE/CVF Conference on Computer Vision and Pattern Recognition (CVPR)}, 
  title={DreamBooth: Fine Tuning Text-to-Image Diffusion Models for Subject-Driven Generation}, 
  year={2023},
  volume={},
  number={},
  pages={22500-22510},
  doi={10.1109/CVPR52729.2023.02155}}

@misc{shipard2023diversity,
      title={Diversity is Definitely Needed: Improving Model-Agnostic Zero-shot Classification via Stable Diffusion}, 
      author={Jordan Shipard and Arnold Wiliem and Kien Nguyen Thanh and Wei Xiang and Clinton Fookes},
      year={2023},
      eprint={2302.03298},
      archivePrefix={arXiv},
      primaryClass={cs.CV},
      url={https://arxiv.org/abs/2302.03298}, 
}

@inproceedings{li2023blip2,
author = {Li, Junnan and Li, Dongxu and Savarese, Silvio and Hoi, Steven},
title = {BLIP-2: bootstrapping language-image pre-training with frozen image encoders and large language models},
year = {2023},
publisher = {JMLR.org},
booktitle = {Proceedings of the 40th International Conference on Machine Learning},
articleno = {814},
numpages = {13},
location = {Honolulu, Hawaii, USA},
series = {ICML'23}
}

@misc{liu2024llava,
      title={Visual Instruction Tuning}, 
      author={Haotian Liu and Chunyuan Li and Qingyang Wu and Yong Jae Lee},
      year={2023},
      eprint={2304.08485},
      archivePrefix={arXiv},
      primaryClass={cs.CV},
      url={https://arxiv.org/abs/2304.08485}, 
}

@misc{wang2024qwen2vl,
      title={Qwen2-VL: Enhancing Vision-Language Model's Perception of the World at Any Resolution}, 
      author={Peng Wang and Shuai Bai and Sinan Tan and Shijie Wang and Zhihao Fan and Jinze Bai and Keqin Chen and Xuejing Liu and Jialin Wang and Wenbin Ge and Yang Fan and Kai Dang and Mengfei Du and Xuancheng Ren and Rui Men and Dayiheng Liu and Chang Zhou and Jingren Zhou and Junyang Lin},
      year={2024},
      eprint={2409.12191},
      archivePrefix={arXiv},
      primaryClass={cs.CV},
      url={https://arxiv.org/abs/2409.12191}, 
}

@misc{carion2025sam3segmentconcepts,
      title={SAM 3: Segment Anything with Concepts},
      author={Nicolas Carion and Laura Gustafson and Yuan-Ting Hu and Shoubhik Debnath and Ronghang Hu and Didac Suris and Chaitanya Ryali and Kalyan Vasudev Alwala and Haitham Khedr and Andrew Huang and Jie Lei and Tengyu Ma and Baishan Guo and Arpit Kalla and Markus Marks and Joseph Greer and Meng Wang and Peize Sun and Roman Rädle and Triantafyllos Afouras and Effrosyni Mavroudi and Katherine Xu and Tsung-Han Wu and Yu Zhou and Liliane Momeni and Rishi Hazra and Shuangrui Ding and Sagar Vaze and Francois Porcher and Feng Li and Siyuan Li and Aishwarya Kamath and Ho Kei Cheng and Piotr Dollár and Nikhila Ravi and Kate Saenko and Pengchuan Zhang and Christoph Feichtenhofer},
      year={2025},
      eprint={2511.16719},
      archivePrefix={arXiv},
      primaryClass={cs.CV},
      url={https://arxiv.org/abs/2511.16719},
}

@misc{labs2025flux1kontextflowmatching,
      title={FLUX.1 Kontext: Flow Matching for In-Context Image Generation and Editing in Latent Space},
      author={Black Forest Labs and Stephen Batifol and Andreas Blattmann and Frederic Boesel and Saksham Consul and Cyril Diagne and Tim Dockhorn and Jack English and Zion English and Patrick Esser and Sumith Kulal and Kyle Lacey and Yam Levi and Cheng Li and Dominik Lorenz and Jonas Müller and Dustin Podell and Robin Rombach and Harry Saini and Axel Sauer and Luke Smith},
      year={2025},
      eprint={2506.15742},
      archivePrefix={arXiv},
      primaryClass={cs.GR},
      url={https://arxiv.org/abs/2506.15742},
}

@misc{fu2023dreamsim,
      title={DreamSim: Learning New Dimensions of Human Visual Similarity using Synthetic Data}, 
      author={Stephanie Fu and Netanel Tamir and Shobhita Sundaram and Lucy Chai and Richard Zhang and Tali Dekel and Phillip Isola},
      year={2023},
      eprint={2306.09344},
      archivePrefix={arXiv},
      primaryClass={cs.CV}
}

@article{wang2025review,
  title   = {Review of Surface-defect Detection Methods for Industrial Products Based on Machine Vision},
  author  = {Fang, Wei and Wang, Mengnan and Sun, Jiadong and Chen, Deji and Shi, Pei},
  journal = {IEEE Access},
  volume  = {13},
  pages   = {90668--90697},
  year    = {2025},
  month   = may,
  doi     = {10.1109/ACCESS.2025.3571297},
}

@misc{bai2025comprehensivesurveymachinelearning,
  title        = {A Comprehensive Survey on Machine Learning Driven Material Defect Detection},
  author       = {Bai, Jun and Wu, Di and Shelley, Tristan and Schubel, Peter and Twine, David and Russell, John and Zeng, Xuesen and Zhang, Ji},
  year         = {2025},
  eprint       = {2406.07880},
  archivePrefix= {arXiv},
  primaryClass = {cs.CV},
  doi          = {10.1145/3730576},
  url          = {https://arxiv.org/abs/2406.07880},
  note         = {Accessed: 2026-01-26},
}

@article{COCO,
  author       = {Tsung{-}Yi Lin and
                  Michael Maire and
                  Serge J. Belongie and
                  Lubomir D. Bourdev and
                  Ross B. Girshick and
                  James Hays and
                  Pietro Perona and
                  Deva Ramanan and
                  Piotr Doll{\'{a}}r and
                  C. Lawrence Zitnick},
  title        = {Microsoft {COCO:} Common Objects in Context},
  journal      = {CoRR},
  volume       = {abs/1405.0312},
  year         = {2014},
  url          = {http://arxiv.org/abs/1405.0312},
  eprinttype   = {arXiv},
  eprint       = {1405.0312},
  bibsource    = {dblp computer science bibliography, https://dblp.org}
}

@misc{sinha20256dstrawberryposeestimation,
AUTHOR = {Sinha, S. N. and K\"uhn, P. J. and Goschke, M. S. and Weinmann, M.},
TITLE = {6D Strawberry Pose Estimation: Real-time and Edge AI Solutions Using Purely Synthetic Training Data},
JOURNAL = {ISPRS Annals of the Photogrammetry, Remote Sensing and Spatial Information Sciences},
VOLUME = {XI-2-2026},
YEAR = {2026},
PAGES = {751--758},
URL = {https://isprs-annals.copernicus.org/articles/XI-2-2026/751/2026/},
DOI = {10.5194/isprs-annals-XI-2-2026-751-2026}
}

\end{document}